\pdfoutput=1
\documentclass[11pt]{article}
\usepackage[]{acl}
\usepackage{times}
\usepackage{latexsym}
\usepackage[T1]{fontenc}
\usepackage[utf8]{inputenc}
\usepackage{microtype}
\usepackage{inconsolata}
\usepackage{graphicx}
\usepackage{booktabs}
\usepackage{svg}
\usepackage{tabularx}
\usepackage{siunitx}
\usepackage{float}
\usepackage{multirow} 
\usepackage{adjustbox} 
\usepackage{makecell} 
\usepackage{enumitem}
\usepackage[colaction]{multicol}
\usepackage{hyperref}
\usepackage{verbatim}
\usepackage{subfiles}
\usepackage{caption}
\usepackage{subcaption}
\usepackage{rotating}
\usepackage{makecell}

\newcolumntype{H}{>{\setbox0=\hbox\bgroup}c<{\egroup}@{}}
\newcolumntype{C}{>{\Centering\arraybackslash}X} % centered "X" column

\newenvironment{Table}
  {\par\bigskip\noindent\minipage{\columnwidth}\centering}
  {\endminipage\par\bigskip}

\title{PeerQA: A Scientific Question Answering Dataset from Peer Reviews}

\author{
Tim Baumgärtner,$^{1}$ 
Ted Briscoe,$^{2}$ 
Iryna Gurevych$^{1,2}$ \\
$^{1}$Ubiquitous Knowledge Processing Lab (UKP Lab), \\ Department of Computer Science and Hessian Center for AI (hessian.AI), \\Technical University of Darmstadt \\
$^{2}$Mohamed bin Zayed University of Artificial Intelligence\\
\url{www.ukp.tu-darmstadt.de}
}

\begin{document}
\maketitle
\begin{abstract}
We present PeerQA, a real-world, scientific, document-level Question Answering (QA) dataset. PeerQA questions have been sourced from peer reviews, which contain questions that reviewers raised while thoroughly examining the scientific article. Answers have been annotated by the original authors of each paper. The dataset contains 579 QA pairs from 208 academic articles, with a majority from ML and NLP, as well as a subset of other scientific communities like Geoscience and Public Health.
PeerQA supports three critical tasks for developing practical QA systems: Evidence retrieval, unanswerable question classification, and answer generation. 
We provide a detailed analysis of the collected dataset and conduct experiments establishing baseline systems for all three tasks. Our experiments and analyses reveal the need for decontextualization in document-level retrieval, where we find that even simple decontextualization approaches consistently improve retrieval performance across architectures. On answer generation, PeerQA serves as a challenging benchmark for long-context modeling, as the papers have an average size of 12k tokens.\footnote{Our code and data is available at \url{https://github.com/UKPLab/peerqa}.}

\end{abstract}

\begin{figure}[!ht]
\centering
\includegraphics[width=0.89\columnwidth]{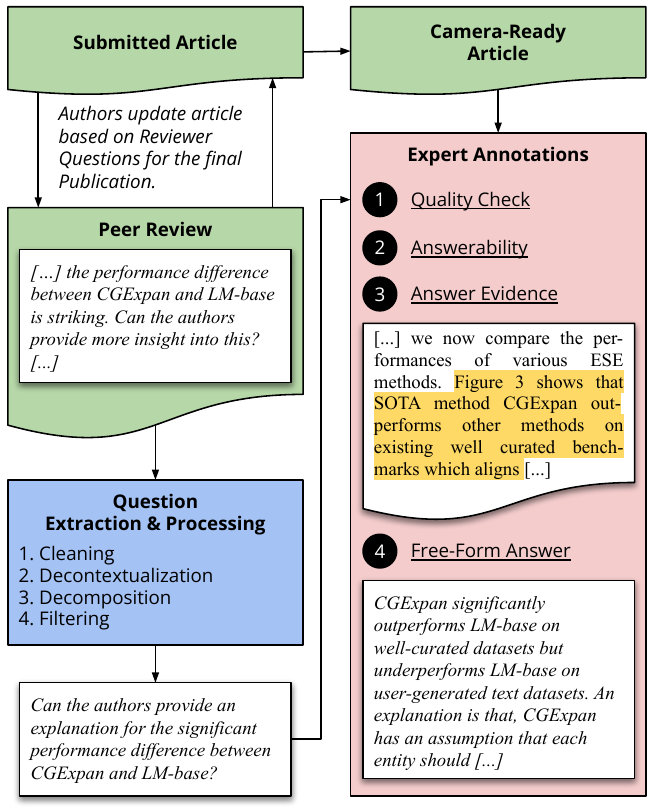}
\caption{Overview of the PeerQA data collection process. From the peer review process (in green), we extract and process questions from the reviews. Given the published version of the article and a question, an expert (in our case, the original paper authors) (1) checks the question and modifies or discards it, (2) annotates whether it is answerable or not (i.e. if there is sufficient information in the paper), and if so (3) highlights the evidence to answer the question and finally (4) provides a free-form answer to the question.}
\label{fig:peer-qa-process}
\end{figure}

\section{Introduction}
The number of scientific articles is increasing exponentially \citep{fire-over-optimization-2019,growth-rate-modern-science}, leading to an increase in review work and leaving researchers with an ever-expanding number of publications to read to keep up with their field. Therefore, novel tools are required to support reviewing work and enable readers to consume information from scientific articles more efficiently \citep{brainard2020scientists,kuznetsov2024natural}.
\begin{table*}
\small
\centering
\begin{tabular}{@{}lllllllll@{}}
\toprule
Dataset & Papers & QA & Domain & \begin{tabular}[c]{@{}l@{}}Questioner\\Knowledge\end{tabular} & \begin{tabular}[c]{@{}l@{}}Question\\Source\end{tabular} & Annotators & \begin{tabular}[l]{@{}l@{}}Answer\\Source\end{tabular} & \begin{tabular}[l]{@{}l@{}}Answer\\Types\end{tabular}  \\ \midrule
BioASQ & 43011 & 4615 & BioMed. & -- & -- & Experts &  Abstract & Y/N, Ex, FF \\
QASPER & 1585 & 5049 & NLP  & Abstract & Crowdsourced & Practitioners &  Paper & Y/N, Ex, FF, U/A \\
QASA & 113 & 1798 & AI/ML & Full Paper & Crowdsourced & Practitioners &  Paper &  Ex, FF, U/A \\ \midrule
PeerQA & 208 & 579 & Multi & Full Paper & Reviews  & Experts &  Paper & Ex, FF, U/A \\ \bottomrule
\end{tabular}
\caption{Comparison of the most relevant scientific QA datasets. In BioASQ, experts come up with questions without a document in mind. Answer types abbreviations: Y/N = Yes/No, Ex = Extractive or Evidence Retrieval, FF = Free-Form Answers, U/A = Unanswerable). The QA column reports the number of question-answer annotations.
\looseness=-1}
\label{tbl:related-work-dataset-comparision}
\end{table*}
Automatic Question Answering (QA) systems can provide such support, allowing researchers and reviewers to productively extract information from an article, particularly if integrated directly into the reading and reviewing interface \citep{zyska-etal-2023-care, lo-semantic-reader-2023}. QA systems can also improve the quality of peer review, e.g., by avoiding questions in a review that are addressed in the article but potentially overlooked by a reviewer. However, the development of QA models is limited by the availability of high-quality and realistic datasets in the scientific domain to measure the performance of methods. Collecting scientific QA data is challenging because it requires expert annotators who are difficult to recruit. Furthermore, naturally occurring questions are difficult to source compared to the general domain, where search engine logs can be used \citep{nguyen-msmarco-2016,kwiatkowski-etal-2019-natural}. Previous work resorted to recruiting practitioners or graduate students and focused only on Machine Learning (ML) or Natural Language Processing (NLP) domains \citep{dasigi-etal-2021-dataset, qasa-lee-2023}. Annotators of these datasets have various degrees of knowledge, e.g., having read only the abstract, skimmed the paper, or sometimes read the paper fully. Collecting questions from annotators has the downside of questions not being realistic, such as asking questions that would not be raised naturally or being generic when the questioner has superficial knowledge of the paper.

To this end, we introduce PeerQA, a real-world, scientific, document-level Question Answering dataset. PeerQA supports three crucial tasks for QA over scientific articles: Given a question and a paper, evidence sentences relevant to the question need to be retrieved. Based on these, the answerability of the question can be decided. Finally, the dataset contains free-form reference answers addressing the question. We leverage peer reviews to source questions, and answers are annotated by the authors of the respective papers. While most questions are from ML and NLP papers, 10\% of questions come from other scientific domains, including Geoscience and Public Health. Figure~\ref{fig:peer-qa-process} provides an overview of our data collection process. To summarize, our contributions are the following:\looseness=-1

1. We release PeerQA, a QA dataset over scientific articles with questions sourced from peer reviews and answers annotated by authors. We release a set of 579 annotated samples (from 208 papers), as well as 12k unlabeled questions (from 2.6k papers). We show the properties of the collected data, including various statistics, question topics, and classes.

2. We establish baselines for all three tasks in PeerQA: Evidence Retrieval, Question Answerability, and Free-Form Answer Generation, and outline which factors contribute to model performance.

\section{Related Work}
\paragraph{Peer Review} Many tasks and applications leverage peer reviews as a data source, including argument mining \citep{hua-etal-2019-argument, cheng-etal-2020-ape, kennard-etal-2022-disapere}, helpfulness and score prediction \citep{xiong-litman-2011-automatically, gao-etal-2019-rebuttal}, review generation \citep{auto-peer-review,darcy-marg-2024}, tagging and linking review comments with the paper \citep{kuznetsov-etal-2022-revise, darcy-etal-2024-aries}, rebuttal generation \citep{purkayastha-etal-2023-exploring}, the study and analysis of peer review \citep{kang-etal-2018-dataset, ghosal2022peer} and more general contexts such as document revision \citep{ruan-etal-2024-re3}. In PeerQA, we utilize peer reviews to source a scientific QA dataset.

\paragraph{Scientific QA} QA datasets in the scientific domain can generally be categorized as larger-scale datasets that are (semi-) automatically created and small expert-annotated datasets. 

\begin{table*}[!ht]
\small
\centering
\begin{tabular}{@{}llcccccc@{}}
\toprule
Venue & Domain & Papers & Questions & Evidence & Free-Form & Ev. \& FF. & Unanswerable \\ \midrule
ICLR 23 & ML & 49 & 153 & 103 & 89 & 63 & 36 \\
ICLR 22 & ML & 44 & 137 & 107 & 75 & 68 & 14 \\
NeurIPS 22 & ML & 25 & 79 & 56 & 51 & 40 & 16 \\ 
ARR 22 & NLP & 45 & 87 & 60 & 61 & 49 & 21 \\
COLING 20 & NLP & 15 & 31 & 25 & 13 & 13 & 5 \\
ACL 17 & NLP & 7 & 20 & 16 & 10 & 9 & 4 \\
CoNLL 16 & NLP & 5 & 12 & 7 & 7 & 7 & 5 \\
ESD 23 & Geoscience & 5 & 17 & 10 & 11 & 5 & 1 \\
ESurf 23 & Geoscience & 3 & 16 & 16 & 9 & 9 & 0 \\
F1000 22 & Mixed & 10 & 27 & 14 & 10 & 4 & 10 \\
\midrule
Total &  & \textbf{208} & \textbf{579} & \textbf{414} & \textbf{336} & \textbf{267} & \textbf{112} \\ \bottomrule
\end{tabular}
\caption{Number of collected question-answer pairs per venue in PeerQA. \textit{Evidence} shows the number of questions that have at least one sentence annotated addressing the question. \textit{Free-Form} reports the number of questions with an annotated free-form answer. The \textit{Ev. \& FF.} column reports the union of both. Finally, the \textit{Unanswerable} column reports the number of questions that can not be answered due to insufficient information in the paper.}\label{tbl:dataset-nums}
\end{table*}

Among the larger-scale but (semi-) automatically created QA datasets are PubMedQA \citep{jin-etal-2019-pubmedqa}, in which questions are sourced from article titles that are phrased as questions. Answers are either yes, no, or maybe, and a subset is expert-annotated. SciDefinition \citep{august-etal-2022-generating} uses templates to generate questions about the definition of scientific terms. \citet{kulshreshtha-etal-2021-back} create a dataset in the ML and Biomedicine domain with questions sourced from Google's "People also ask" suggestions and answers from the search engine's span extraction feature. 
\citet{wan2024sciqag} generate a large-scale, scientific QA dataset by distilling a generation model from \mbox{GPT-4} instructed to output QA pairs given a paper. 
\citet{auer-et-al-2023-sciqa} develop question templates to automatically generate questions that are answerable from the Open Research Knowledge Graph \citep{jaradeh-et-al-2019-orkg} covering factoid questions, e.g., about the metadata of a paper, or questions that require inference over multiple papers. The questions in PeerQA are all focused on a single publication and the content of it, and our baselines use only the unstructured text of the article. PeerQA is an expert-annotated QA resource, where questions are sourced from human-written peer reviews and answers are annotated by paper authors.\looseness=-1

Regarding expert annotated datasets, the BioASQ challenge \citep{bioasq,krithara-et-al-2022-bioasq-qa} is an open-domain QA dataset from biomedical experts. Experts come up with questions and corresponding answers (yes/no, factoid, list, and free-form), which are additionally grounded in sentences from abstracts of publications on PubMed. While this is one of the greatest available resources for biomedical QA, annotating answers only in abstracts limits the question and answer complexity. Compared with PeerQA, questions are also more general, i.e., they are not asked within the context of a specific paper, and answers can be found in various articles.
Most similar to our work are the QASPER \citep{dasigi-etal-2021-dataset} and QASA \citep{qasa-lee-2023} datasets. In QASPER, NLP practitioners have read the abstract of a paper and raised questions about the paper. This leads to generic questions applicable to many papers (e.g., "Which baselines did they compare?") and questions that are easy to answer from the full paper. QASA takes this a step further by giving question annotators access to the full paper, instructing them to either skim or read it in more detail. In both these datasets, annotators create questions and answers; in contrast, our questions are based on peer reviews, i.e., they have been naturally raised by a reviewer, a domain expert who has read the paper in detail. Besides the questions, the answers in PeerQA are provided by experts, i.e., the authors of the respective papers. Table~\ref{tbl:related-work-dataset-comparision} provides an overview of these differences. To summarize, PeerQA is the first scientific QA resource with natural questions and all QA pairs annotated by paper authors.

In concurrent work, \citet{singh-etal-2024-scidqa} also explore extracting questions from peer reviews in the ML domain. Unlike PeerQA, their approach uses the authors' responses provided during the rebuttal to obtain reference answers. To identify supporting evidence from the paper for each answer, they employ a hybrid approach that combines manual and automated mapping of the answers to relevant information in the paper.

\begin{figure*}
    \centering
    \includegraphics[width=\textwidth]{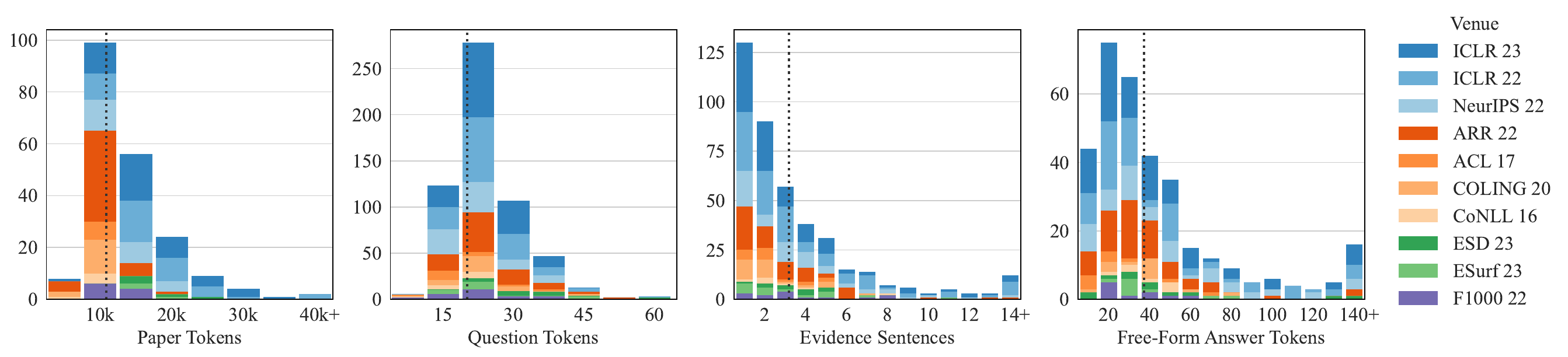}
    \caption{Statistics of the PeerQA dataset. The color coding shows the distribution per venue and by the scientific community (i.e., blue colors for ML, orange for NLP, green for Geosciences, and purple for mixed). The gray dotted line indicates the average. The leftmost histogram shows a paper distribution, while the others show a distribution of questions. We measure the number of tokens using the \texttt{Llama-3} tokenizer.}
    \label{fig:dataset-stats}
\end{figure*}

\paragraph{Long-Context QA} Dialogue and QA systems grounded in a document have recently gained traction \citep{dialdoc-2023-dialdoc}. 
In this vein, NarrativeQA \citep{kocisky-etal-2018-narrativeqa} contains questions about movie scripts and books with an average length of 63k tokens. \citet{pang-etal-2022-quality} construct a multiple-choice dataset over books and articles with an average length of 5k tokens focusing on questions that require reading the article in detail. \citet{reddy-etal-2024-docfinqa} extend FinQA \citep{chen-etal-2021-finqa} to financial documents with an average of 123k words. ConditionalQA~\citep{sun-etal-2022-conditionalqa} is a dataset of government documents with an average length of 1.5k tokens and answers tied to certain input conditions. PeerQA serves as another resource for long-context QA, with documents having an average length of 12k tokens and 30\% of questions requiring combining information from more than one location in the paper.

\section{PeerQA}
\subsection{Data Collection}\label{sec:data-collection} 
Figure~\ref{fig:peer-qa-process} provides an overview of the data collection process. We use papers and peer reviews from NLPeer \citep{dycke-etal-2023-nlpeer} and extend this set with journals and conferences that publish peer reviews and camera-ready versions publicly. Specifically, the data from ARR 2022 (containing papers published at ACL and NAACL 2022), COLING 2020, ACL 2017, CoNLL 2016, and F1000 was curated in NLPeer, partially based on previous data collections \citep{kang-etal-2018-dataset, kuznetsov-etal-2022-revise} and published under a CC-BY-NC-SA 4.0 license. The data from the Geoscience domain is published under a CC-BY 4.0 license in two journals: Earth System Dynamics\footnote{\url{https://www.earth-system-dynamics.net}} (ESD) and Earth Surface Dynamics\footnote{\url{https://www.earth-surface-dynamics.net}} (ESurf). For ICLR 2022/2023 and NeurIPS Datasets and Benchmark Track 2022, we retrieve papers and reviews from OpenReview. Since they are without any license, we do not publish them in our release but provide a download and processing script. All questions and answers in PeerQA are published under CC-BY-NC-SA 4.0.

\paragraph{Paper Processing.} We extract the full text of the camera-ready version of a publication, including equations and captions, using GROBID~0.8~\citep{GROBID}, which also groups sentences into paragraphs, which we use later in our experiments.

\paragraph{Question Processing.} From the peer reviews of each paper, we extract an initial set of questions using all sentences ending in a question mark, resulting in 17910 questions.\footnote{In preliminary experiments, we extracted questions based on syntax. However, this resulted in many false positives.} The resulting questions comprise three problems: First, they are noisy as peer reviews often contain spelling or grammar mistakes. Second, they are contextualized into the preceding sentences of the review, i.e., their actual meaning can only be understood from the context of the review but not in isolation. Third, some questions contain compounds of multiple or follow-up questions after applying the decontextualization step. We deemed this problematic for our annotations as it would obfuscate which evidence aligns with which part of the question. To address these issues, we conduct two preprocessing steps: First, we create a clean and contextualized version of a question using InstructGPT\footnote{We use \texttt{text-davinci-003}. However, when we added the Geoscience subset, \texttt{text-davinci-003} was no longer available. Thus, we resorted to \texttt{gpt-4-0125-preview}.} \citep{ouyang-instruction-2022}. For this, we prompt the model with the preceding three sentences of the review and the extracted question to generate a single question that is context-independent. Conveniently, due to the good fluency of Large Language Models (LLM), this also addresses the noisiness of the original question. To detect multiple or follow-up questions, we employ a constituency parser \citep{kitaev-klein-2018-constituency, kitaev-etal-2019-multilingual} and flag questions with root-level conjunctions. We then decompose these questions adopting InstructGPT again.

Finally, we manually filter all resulting questions to include only information-seeking types of questions and discard questions that contain errors due to the preprocessing steps or not being relevant for a QA dataset. Specifically, we ensure that questions address the \emph{content} of the paper (e.g., we discard questions of rhetorical nature or about typos and layout) and are \textit{decontextualized} correctly (i.e., we discard questions that are ambiguous, contain hallucinations or references such as line numbers that are not present in a camera-ready version).\footnote{This filtering step has largely been done by a graduate NLP student supported by the paper authors.} In this step, we remove 30\% of the questions, yielding the final set of 12546 questions.

\paragraph{Answer Annotation.} 
Our questions were asked based on the submitted article. However, answers are annotated in the final publication. Hence, our annotation process relies on authors incorporating reviews into the final version for questions to be answerable. Questions might also be answerable when reviewers overlooked details in the submission that already answer their questions.
For each paper, we contact paper authors via email requesting their voluntary participation in answering the questions (see \S\ref{sec:appendix-contact-email}).\footnote{For the 5 CoNLL papers, we were unsuccessful in contacting the authors. Therefore, the annotations were performed by a senior NLP professor and co-author of this paper.} We implement multiple layers to instruct authors on how to complete the task. First, we provide a high-level description of the task in the initial email and a link to the detailed annotation guideline. We updated the annotation guideline during data collection with common questions we received. Moreover, we explain the annotation interface and demonstrate the task in a short video. Finally, our annotation interface (see \S\ref{sec:appendix-annotation-interface}) also contains UI elements that provide hints to the authors explaining the task. The annotation task comprises 4 steps: First, authors can provide feedback on a question, e.g., to remove or update it. Second, the authors highlight any text in the PDF of the final paper that is relevant to answering the question, which we refer to as Answer Evidence. Third, the authors provide free-form text that directly answers the question. Alternatively, questions can also be flagged as unanswerable. Unanswerable questions can, for example, occur when a question from a reviewer has been answered in the rebuttal but was not incorporated into the final publication. While we ask authors to perform all steps, some questions only have answer evidence or a free-form answer, but not both. The annotated evidence is mapped to the text extracted from the PDF. We notice that GROBID occasionally misses paragraphs that can not be mapped to the annotated evidence. We publish the raw annotated data and the mapped data, allowing future research with access to better PDF extraction tools to use the full dataset.\looseness=-1

\paragraph{Quality Control} Besides manually filtering questions and removing low-quality or irrelevant ones, we also provide the experts with a way to improve the dataset's quality. In our annotation interface, authors can leave feedback for a question, e.g., if they find it imprecise and wish to correct or remove it. All feedback has been manually processed, and the questions have been updated or removed.
Finally, we notice a high variance in the free-form answer quality. While some answers are clear and concise, others are more succinct and provide less detail. Although we give detailed guidelines on how to write the free-form answer to the authors, since we only engage briefly with them, it is challenging to enforce a similar quality. To counter this, we augment the collected answers with rephrases from GPT-4 \citep{openai-gpt-4}.\footnote{See \S\ref{sec:appendix-answer-augmentation-with-evidence} and \S\ref{sec:appendix-answer-augmentation-without-evidence} for prompts.}

Following this process, we obtained 579 answers from 208 papers. Table~\ref{tbl:dataset-nums} reports the number of annotations per venue. We also release the remaining 11967 questions from 2623 papers that have not been answered.\footnote{The number of mapped evidence from the noisy text extraction is reported in \S\ref{sec:appendix-dataset-numbers-with-extraxted}. Examples are provided in \S\ref{sec:appendix-examplary-annotation}. \S\ref{sec:appendix-unlabeled-data} reports a breakdown by venue for the unlabeled questions.}

\subsection{Analysis}\label{sec:analysis}
We report distributional statistics of the dataset in Figure~\ref{fig:dataset-stats}. Notably, the average paper length is 11723 tokens, which provides an interesting benchmark for long-context generative models. Furthermore, questions are relatively long, with an average of $20.2$ tokens (the average length in BioASQ, QASPER, and QASA is $13.2$, $10.2$, and $17.7$, respectively). One reason for this is the question processing pipeline, particularly the decontextualization step. Reviewers construct questions potentially consisting of multiple sentences. During preprocessing, the question has been rephrased to contain all this information. We analyze the semantic similarity between the final and original questions, finding that $90\%$ of questions have a similarity of more than $0.6$ and $50\%$ more than $0.82$.\footnote{\S\ref{sec:cosine-similarity-question-context} provides a detailed analysis of the similarities.} This shows that our processed questions remain highly similar to the original questions in the review. On average, questions have 3.8 annotated answer evidence sentences. Besides, 30\% of questions have non-consecutive answer evidence, i.e., the evidence is distributed non-contiguously over the paper.\footnote{\S\ref{sec:appendix-answer-evidence-stats} reports more answer evidence statistics.}\looseness=-1

\begin{table*}[htpb]
\small
\centering
\begin{tabular}{@{}llcccccccc@{}}
\toprule
 &  & \multicolumn{4}{c}{MRR} & \multicolumn{4}{c}{Recall@10} \\ \cmidrule(lr){3-6}
\cmidrule(lr){7-10}
 Model & Architecture & Para. & +Title & Sent. & +Title & Para. & +Title & Sent. & +Title \\ \midrule
MiniLM-L12-v2 & Cross-Encoder & \textbf{0.4723} & \underline{0.4839} & \textbf{0.3644} & \textbf{0.3654} & 0.6467 & 0.6709 & 0.3505 & \textbf{0.3746} \\
Contriever & Dense & 0.3494 & 0.3624 & 0.2778 & 0.2773 & 0.5567 & 0.5340 & 0.2896 & 0.2910 \\
Contriever-MS & Dense & 0.4095 & 0.4408 & 0.3184 & 0.3160 & 0.6160 & 0.6314 & 0.3361 & 0.3538 \\
Dragon+ & Dense & \underline{0.4657} & \textbf{0.4845} & 0.3345 & 0.3433 & 0.6563 & \underline{0.6817} & \underline{0.3637} & 0.3667 \\
GTR-XL & Dense & 0.3955 & 0.4142 & 0.3048 & 0.2981 & 0.5940 & 0.6122 & 0.3522 & 0.3190 \\
ColBERTv2 & Multi-Dense & 0.4368 & 0.4122 & \underline{0.3480} & \underline{0.3491} & 0.6287 & 0.6371 & 0.3607 & 0.3544 \\
BM25 & Sparse & 0.4288 & -- & 0.2850 & -- & 0.6388 & -- & 0.3058 & -- \\
SPLADEv3 & Sparse & 0.4536 & 0.4725 & 0.3477 & 0.3419 & \textbf{0.6661} & \textbf{0.6851} & \textbf{0.3757} & \underline{0.3687} \\
\bottomrule
\end{tabular}
\caption{Answer evidence retrieval results on paragraph (Para.) and sentence (Sent.) level and with decontextualizing the passages by prepending the title (+Title). Top-scoring models are in bold, and runner-ups are underlined.}
\label{tbl:evidnce-retrieval-results}
\end{table*}

We run a topic model to understand which questions are contained in PeerQA, specifically BERTopic \citep{grootendorst-bertopic-2022}. We find community-specific clusters (e.g., mentions of \textit{language} or \textit{annotation} for NLP; \textit{carbon} or \textit{soil} for Geoscience), topics about specific elements of the paper (e.g., figures, tables, or equations) or specialized clusters (e.g., adversarial attacks or fine-tuning/hyperparameter related questions).\footnote{A list of topics and their size can be found in \S\ref{sec:appendix-question-topcis}. We also apply the topic model to the unlabeled questions.} While the topic analysis clusters questions semantically, we also sample 100 questions randomly and sort them into one of 8 question classes: Methods, Data, Implications, Definitions, Comparisons, Analysis, Justification, and Evaluation.\footnote{The annotation was performed by two graduate students, reaching a substantial agreement of $0.68$ Cohens Kappa.} We find that 44\% of questions aim to clarify the methods or data, followed by 12\% of questions asking the authors to justify a decision.\footnote{Class definitions and the distribution can be found in \S\ref{sec:appendix-question-classes}.}

\section{Experiments}
\subsection{Answer Evidence Retrieval}\label{sec:experiments-answer-evidence-retrieval}
We set up the answer evidence retrieval task as an information retrieval problem: Given a query, the model computes a score for each passage in the paper, where a passage can be a paragraph or sentence. To evaluate the answer evidence retrieval task, we test models of various architectures, including cross-encoder \citep{nogueira-bert-reranking-2019}, dense retrieval \citep{reimers-gurevych-2019-sentence}, multi-vector dense retrieval \citep{khattab-colbert-2020}, sparse \citep{zamani-snrm-2018} and lexical models. Specifically, a cross-encoder model concatenates the query and passage and outputs a relevance score. In contrast, dense approaches encode query and passage independently by the same or individual models, obtaining a high-dimensional representation for each. A score is computed via dot-product or cosine-similarity between the two representations. Multi-vector approaches represent a query and passage not by a single but by many representations, e.g., for each token. The relevance score is computed by taking the sum of the maximum score between each query and passage token. Lexical approaches use term matching and weighting between the query and passage. Building upon this, sparse models perform a semantic query and/or document expansion to overcome the lexical gap. Concretely, we evaluate: \texttt{MiniLM-L12-v2} \citep{minilm, beir}, \texttt{Contriever} \citep{izacard-contriever-2022}, \texttt{Dragon+} \citep{lin-etal-2023-train}, \texttt{GTR} \citep{ni-etal-2022-large}, \texttt{ColBERTv2} \citep{santhanam-etal-2022-colbertv2}, \texttt{BM25} \citep{robertson-bm25-2009} and \texttt{SPLADEv3} \citep{lassance-spladev3-2024}.\looseness=-1

Besides various models, we investigate the impact of retrieving paragraphs or sentences. We use the paragraphs extracted by GROBID and mark any paragraph as relevant that contains a relevant sentence. Furthermore, we investigate a baseline to improve the decontextualization by prepending the title, which has been shown beneficial in cases where decontextualization is required \citep{wang-etal-2024-dapr}.\looseness=-1

We evaluate using Mean Reciprocal Rank (MRR) \citep{craswell-mrr-2009}, which considers the first relevant passage in a ranked list. While a typical question in PeerQA often has multiple answer evidence sentences (cf. Figure~\ref{fig:dataset-stats}), they frequently belong to the same paragraph or are close to each other. Therefore, pointing a user to the respective paragraph in a real-world application would already be useful as further relevant information usually clusters around the same location. We also measure the quality of the entire ranking by evaluating Recall@10. We chose 10, as most questions have fewer relevant sentences (cf. Figure~\ref{fig:answer-evidence-stats}).

\begin{figure*}[!t]
    \centering
    \includegraphics[width=\textwidth]{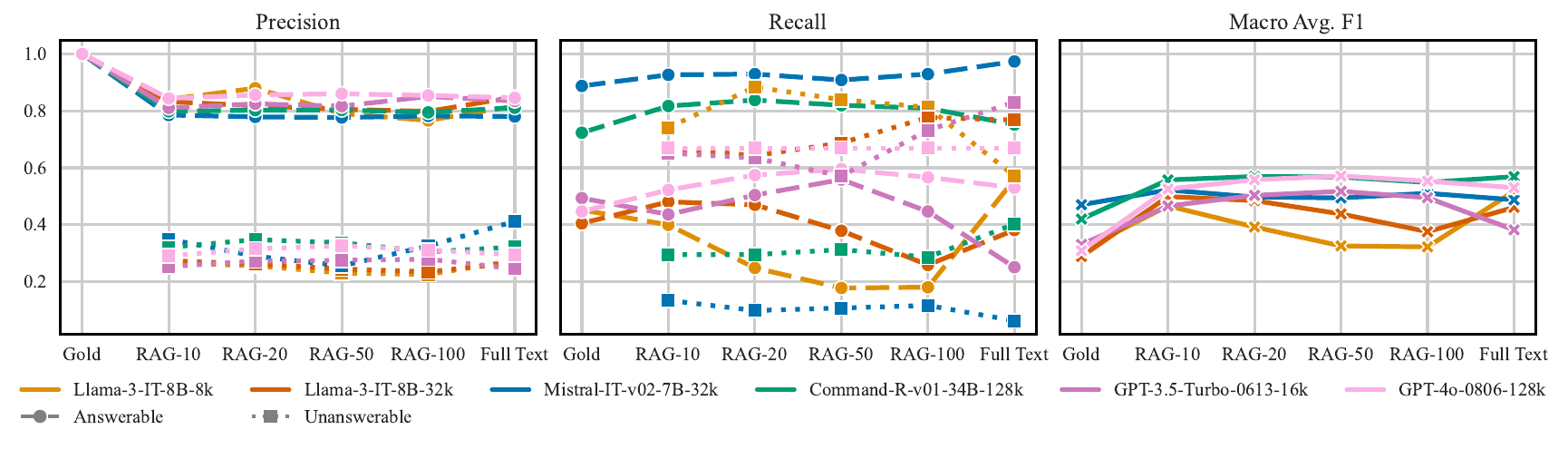}
    \captionof{figure}{Answerability scores (y-axis) with different contexts (x-axis). In the \textit{Gold} setting, the model is only provided with the annotated, relevant paragraphs (i.e., no unanswerable questions are available in this setting); in \textit{Full Text}, the entire paper is provided in the context (and potentially truncated); otherwise, the top-scoring passages by \texttt{SPLADEv3} are provided. The Precision and Recall plots show the Answerable (- -) and Unanswerable ($\cdot\cdot$) classes.}
    \label{fig:results-un-answerable}
\end{figure*}

\subsection{Answerability and Answer Generation}\label{sec:experiments-answerability-answer-generation}
We set up the answerability task as a binary classification problem: given a question and context, a model predicts whether a question is answerable or not. We label all questions as answerable with annotated answer evidence and all as unanswerable, which the authors flagged as such. The answer generation task is set up as a sequence-to-sequence task, i.e., given the question and the context, the answer needs to be generated. For both tasks, we employ instruction-tuned LLMs. For the answerability task, we prompt the model to either answer the question if sufficient evidence is provided or to generate \textit{No Answer}. However, to obtain generations for all answerable questions, we remove the instruction to generate \textit{No Answer} from the prompt for the answer generation task (see \S\ref{sec:appendix-answerability-prompt} and \S\ref{sec:appendix-answer-generation-prompt} for the prompts). We experiment with providing as context the gold passages (ablating retrieval errors), the top-$k$ retrieved paragraphs (where $k \in \{10, 20, 50, 100\}$), and the full text. This is a Retrieval Augmented Generation (RAG) \citep{lewis-rag-2020} setup, except we retrieve from a single, long document instead of a corpus.We truncate the paragraphs if required by the maximum context size of the models and decode greedily from the models. Specifically, we use \texttt{Llama-3-8B-Instruct} \citep{llama3modelcard}, which we also extend to a 32k context size with dynamic rope-scaling, \texttt{Command-R}\footnote{\url{https://docs.cohere.com/docs/command-r}}, \texttt{Mistral-7B-Instruct-v0.2} \citep{jiang-mistral-2023}, \texttt{GPT-3.5-Turbo-0613-16k} and \texttt{GPT-4o-0806} \citep{openai-gpt-4}. We evaluate the answerability task as a binary classification problem. We evaluate with macro-F1 to counter the imbalance between the number of answerable (383) and unanswerable (112) questions.

Evaluating generative AI for long-form QA is a challenging, ongoing research topic by itself \citep{krishna-etal-2021-hurdles, xu-etal-2023-critical}. We chose a diverse set of evaluation metrics, including \mbox{Rouge-L} \citep{lin-2004-rouge}, AlignScore \citep{zha-etal-2023-alignscore} and Prometheus-2 \citep{kim-etal-2024-prometheus}. AlignScore is a model-based metric trained on a broad range of text alignment data, among others, on QA. AlignScore breaks the reference into passages of roughly 350 words and the generation into sentences. The model is trained to measure how much each generated sentence is aligned with the information in the reference passage. In practice, we notice that free-form answers provided by the authors can contain information that is not present in the paper. Therefore, besides using only the free-form answer as ground truth, we also compare the generation to the concatenated answer evidence paragraphs. The Prometheus-2 model is an LLM-as-a-judge model \citep{zheng2023judging} fine-tuned on feedback and judgment data generated by GPT-4 on a large set of custom score rubrics. We provide a scoring rubric that measures the correctness of the generated answer given the reference on a scale from 1-5.\footnote{See \S\ref{sec:appendix-evaluation-metric-details} for the Prometheus prompt and score rubric.}\looseness=-1

\begin{figure*}[!t]
    \centering
    \includegraphics[width=1\textwidth]{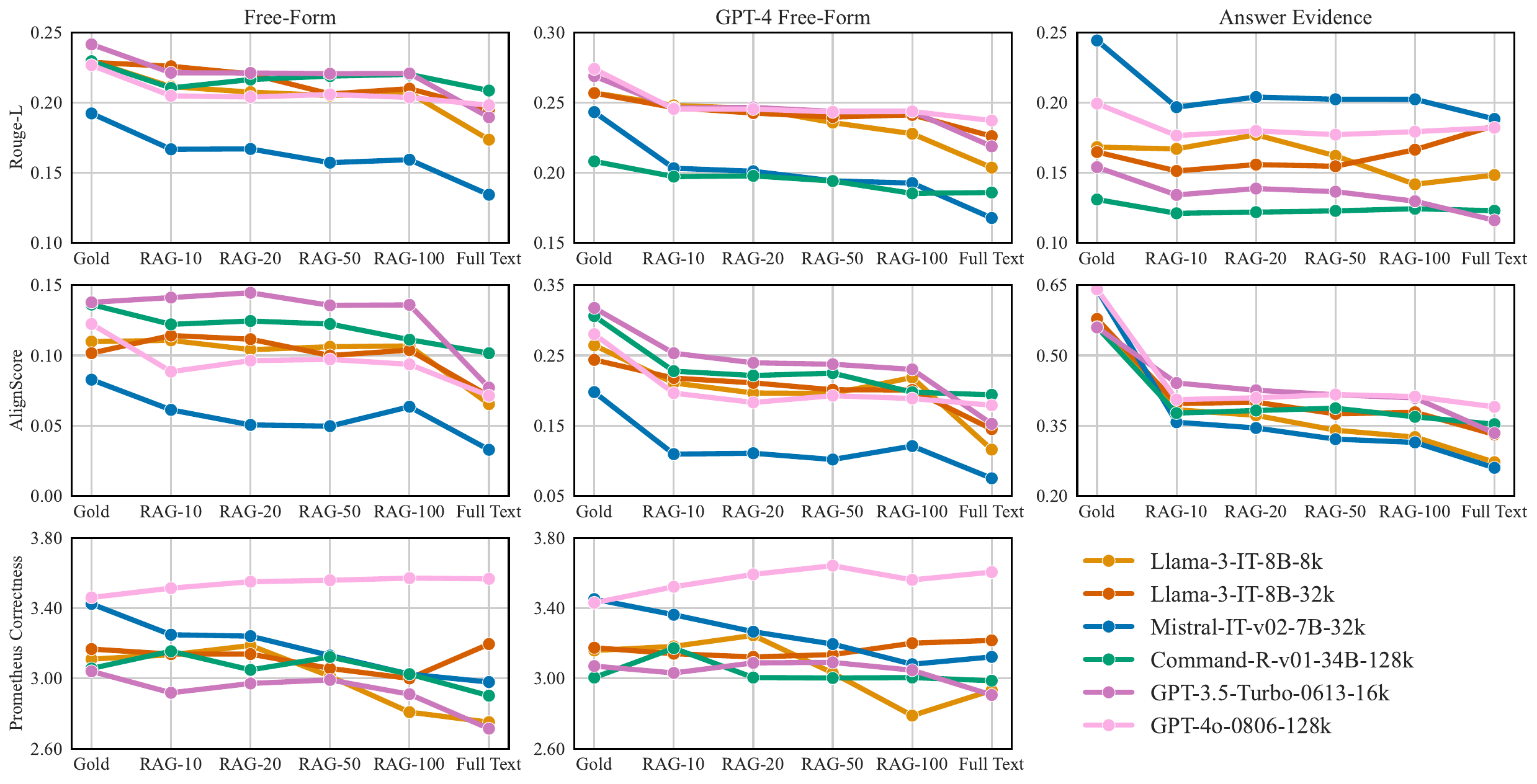}
    \caption{Rouge-L F1, AlignScore and Prometheus Correctness metrics between 
    the annotated free-form answer (1. column), the GPT-4 augmented answer (2. column), the annotated evidence passages (3. column), and the generated answer.\looseness=-1}
    \label{fig:answer-generation-eval}
\end{figure*}

\section{Results}
\subsection{Answer Evidence Retrieval}
Table~\ref{tbl:evidnce-retrieval-results} reports the retrieval results. Across models, we find that retrieving the paragraph yields higher scores than the sentence. Appending the title to the paragraph further improves results (except \texttt{ColBERTv2}'s MRR), showing that decontextualizing the paragraphs from the paper helps. However, we find that MRR and Recall remain the same for most models when prepending the title on a sentence level. Since sentences are short, we conjecture that adding a title influences the overall representation too much, while on a paragraph level, the title only accounts for a fraction of the overall tokens. Overall, we find that \texttt{MiniLM-L12-v2}, \texttt{Dragon+}, and \texttt{SPLADEv3} perform the best. We proceed with \texttt{SPLADEv3} for the RAG experiments as it achieves the highest recall.

\subsection{Answerability}\label{sec:answerability-results}
We report precision, recall, and macro-F1 on the answerability task in Figure~\ref{fig:results-un-answerable}. We observe similar precision for all model and context settings. Precision for answerable questions is much higher than for unanswerable ones. When looking at recall, we find notable differences between the models. While \texttt{Mistral} and \texttt{Command-R} obtain relatively high recall on answerable questions and low recall on unanswerable questions, the \texttt{Llama} and \texttt{GPT} models obtain high recall on unanswerable questions and lower recall on answerable questions. This pattern can be explained: \texttt{Mistral} and \texttt{Command-R} tend to predict an answer more often, while \texttt{Llama} and \texttt{GPT} tend to predict the question is unanswerable, showing that all models have a bias towards one of the classes. \texttt{Command-R} and \texttt{GPT-4o} provide the best trade-off, shown by the highest macro-F1.

\subsection{Answer Generation}\label{sec:results-answer-generation}

Figure~\ref{fig:answer-generation-eval} reports the evaluation metrics comparing the generated answers to either the free-form reference answer, the GPT-4 augmented answer, or the gold paragraphs. Generally, models perform best with the gold answer evidence. Therefore, the annotated evidence provides a strong signal to answer the question. The scores achieved with the gold evidence represent an upper bound. However, higher scores might be possible with more context to better understand the gold answer evidence (or potentially unannotated but useful passages). Upon manual inspection of model errors, we find that lower performance is caused by evaluation failures or information in the free-form answer that is not supported by the evidence (i.e., information that is coming from the author's knowledge that might be general about the field or specific to the paper and did not make it into the camera-ready version). Generally, LLMs perform better in RAG, with fewer but relevant contexts, than in the full-text setting on PeerQA. This shows that despite LLMs' large context sizes, it is more effective to employ a retriever filtering relevant information than leaving this step to the internal workings of the LLM. A notable exception is GPT-4o, which exhibits stable performance with increasing context sizes and increasing performance on answer correctness. GPT-4o is also the most recent and powerful model in our evaluation, demonstrating the improved abilities of state-of-the-art models on long-context tasks. We further analyze the answer generation performance of the RAG setting by measuring the correlation between the retriever recall and the generation metric. We find mostly positive correlations between the retrieval and generation performance across models. While the correlation is not very strong (up to $r=0.42$), it confirms that with increased retrieval performance, the generation improves \cite{salemi-erag-2024}.\footnote{\S\ref{sec:appendix-correlation-generation-recall} reports correlations across all metrics and contexts.}

\begin{table}[!t]
\small
\centering
\begin{tabularx}{0.75\columnwidth}{lS[table-format=3.2]}
\toprule
{Error Class} & \\
\midrule
Correct / Evaluation Error & 43.75\% \\
Partially Correct & 12.50\% \\
Reasoning Error & 10.00\% \\
Implicit Evidence Only & 7.50\% \\
Insufficient Context & 11.25\% \\
Insufficient Evidence & 12.50\% \\
Insufficient Free-Form Answer & 3.75\% \\
\bottomrule
\end{tabularx}
\caption{Error analysis of \texttt{GPT-3.5}'s generations with gold evidence. \S\ref{sec:appendix-error-analysis} provides definitions and examples for error classes.}
\label{tbl:error-analysis}
\end{table}

\paragraph{Error Analysis.} We analyze the lowest performing 80 generations\footnote{Specifically, we sort by the minimum performance of all metrics, considering all questions that have both evidence and free-form annotations and use the \textit{gold} evidence as context.} of \texttt{GPT-3.5} to better understand the errors and report them in Table~\ref{tbl:error-analysis}. We find many low-scoring generations are correct despite at least one of the evaluation metrics providing a low score, for example, when the generation is more verbose or expresses the correct answer differently (\textit{Evaluation Error}). However, we find the metric with the highest score for these generations to be above the 50th percentile in 91\% of the cases. This shows that using different metrics against different ground truths is plausible and catches the alleged failures. Further, we observe the model is only \textit{partially correct} when the free-form answer contains important additional details. In other cases, the model fails to reason correctly over the evidence, e.g., it arrives at an opposite conclusion than the correct answer. Similarly, when the evidence is only implicit or requires expert domain knowledge, the model fails.
Lastly, there are also a few errors in the data. In $11.25\%$ of cases, the gold evidence is not self-sufficient, i.e., more context from the paper would be required, e.g., to understand previously introduced concepts. These errors can likely be recovered through additional retrieval. Other times the answer by the authors is not entailed by the evidence (\textit{Insufficient Evidence}) or the free-form answer only reports the element in the article, but not an actual answer (\textit{Insufficient Free-Form Answer}).\looseness=-1

\section{Conclusion}
We introduced the PeerQA dataset to advance and study question answering on scientific documents. We sourced PeerQA's questions from peer reviews and obtained answer annotations from the paper authors. Our dataset supports three crucial tasks for developing QA systems: evidence retrieval, answerability, and answer generation. We analyzed the collected data and established baseline systems for all three tasks. For evidence retrieval, we find that decontextualization is key to improving performance. On the answerability task, we find that models tend to either over- or under-answer, showing a bias for one of the classes. Further, although models can fit the entire paper into context in the answer generation task, providing the model with the top passages from a retriever outperforms the full-text setting. We also show that with increased retrieval performance, the answer generation improves. Finally, our error analysis highlights the need for better evaluation metrics and model reasoning abilities.\looseness=-1

\section{Limitations}
\paragraph{Dataset Size.} General domain QA datasets usually comprise up to three magnitudes more data than PeerQA (e.g., NQ has 323k samples). However, collecting high-quality data in the scientific domain is challenging due to the requirement for expert annotators. Since science has many domains, it is impractical to collect training data for each of them. Instead, models need to generalize in an unsupervised manner, at most leveraging few-shot examples. Therefore, we introduce PeerQA as an evaluation resource to test the generalizability of models. The size is in line with other recent datasets such as HumanEval \citep{human-eval-chen-et-al-2021} (164 examples), TruthfulQA \citep{lin-etal-2022-truthfulqa} (817), and GPQA \citep{gpqa-rein-et-al-2023} (448). In addition, we release the unlabeled data, comprising 12k questions from 2.6k papers, that can be used for more annotations, unsupervised learning, and further study of reviews. Small evaluation datasets also have the advantage of reduced iteration time over experimental settings, lesser use of compute resources, and a smaller environmental impact.\looseness=-1

\paragraph{Science Domains.}While PeerQA covers more scientific domains compared to prior work, there is a limited amount of data beyond the ML and NLP domains. A major challenge in data collection is the availability of public peer reviews with openly licensed scientific articles \citep{dycke-etal-2022-yes}. We call on the scientific community to further transform reviewing practices to an open format.

\paragraph{English-Only.}PeerQA is limited to English since it is dominant in scientific writing. Nevertheless, publications in other languages exist, and our data collection framework can be applied to any language. The evaluated retrievers are English-only models (except \texttt{BM25}, which is language-agnostic). Some retrieval models have multi-lingual counterparts (e.g., \texttt{mContriever}); however, due to a lack of multi-language data, their performance remains unclear. Some of the evaluated generative models are also multilingual; the performance in other languages is likely to be different than in English.

\paragraph{Free-Form Annotations.}While authors possess the ultimate expertise in their papers, they usually have knowledge beyond the information in their publications. Some free-form answers contain information not included in the answer evidence. For this reason, we also compare the generated answer with the annotated answer evidence, measuring if the model can produce answers entailed by the information in the paper.

\paragraph{Long-Form QA Evaluation.}Evaluating free-form answers is challenging and an ongoing area of research. To evaluate different aspects, we use three metrics against two ground truths. Ideally, we would have multiple free-form answer references; however, even collecting a single response has proven to be challenging. In the hope of better metrics, we also publish the generated answers of our baselines to facilitate adaptation to future, improved methods.

\paragraph{Methods.}Many LLMs and methods \citep{zhao-llm-survey-2023} exist that could be applied to the tasks in PeerQA. Therefore, more sophisticated and specialized methods might exceed the reported performances. However, we focus on introducing the dataset and establishing baseline systems with widely used retrieval and generative models.

\section{Ethical Considerations}
All annotators in PeerQA are authors of accepted articles at conferences or in journals. We do not collect any of their personal information or who has provided the answers. By the nature of our data collection protocol, we only contact authors who have provided their email publicly along with their publication and contact each author individually. Authors have participated voluntarily in the data collection, and we try to keep their workload low by only asking few questions (on average $2.8$). Furthermore, the authors have largely already answered questions during peer review (see \S\ref{sec:cosine-similarity-question-context}), making them familiar with the questions and answers, further reducing their workload.

One objective of PeerQA is to advance the study of peer review, including developing methods and tools to facilitate the authoring and reviewing of scientific articles. Particularly, LLMs have the potential to support authors and reviewers in their work \citep{kuznetsov2024natural}. However, these models also have biases and weaknesses. For example, in our question answerability task, we clearly observe that some models are biased towards one class, i.e., predicting the question as answerable or unanswerable (see \S\ref{sec:answerability-results}). Therefore, these methods can only be used as assistants that support humans. PeerQA sheds light on these issues, raising awareness of potential weaknesses in these models and their careful application in science.

\section*{Acknowledgements}
This work has been funded by the German Research Foundation (DFG) as part of the QASciInf project (grant GU 798/18-3). Further, we gratefully acknowledge the support of Microsoft with a grant for access to the OpenAI GPT models via the Azure cloud (Accelerate Foundation Model Academic Research).

We thank the anonymous reviewers for their helpful suggestions for improving this paper and Sukannya Purkayastha, Max Eichler, and Haritz Puerto for their insightful feedback throughout the paper-writing process. Our gratitude also goes to Maike Nowatzki for reviewing the questions in the Geoscience domain, to Richard Eckart de Castilho for his support with our annotation platform, and to Sebastian Alles for his assistance in establishing the compute and annotation infrastructure. Finally, we are grateful to all authors who have voluntarily participated in creating this dataset.

\bibliography{custom,anthology}

\clearpage

\appendix

\section{PDF extraction of Answer Evidence}\label{sec:appendix-dataset-numbers-with-extraxted}
Text extraction from PDF is not a perfect process. Unfortunately, this means that some annotated answer evidence (and therefore also answerable questions) must be discarded in our experiments since their evidence has not been extracted correctly. Table~\ref{tbl:dataset-numbers-with-extraxted} shows the number of annotated answer evidence (Evidence), as well as the number of questions whose evidence has been extracted correctly (Evidence Mapped). We nevertheless make the complete dataset public so future research with better PDF processing tools can leverage more annotations.
\begin{table}[!h]
\small
\centering
\begin{tabular}{@{}lccc@{}}
\toprule
Venue & Questions & Evidence & Ev. Mapped \\ \midrule

ICLR 23 & 153 & 103 & 93 \\
ICLR 22 & 137 & 108 & 99 \\
NeurIPS 22 & 79 & 56 & 55 \\ 
ARR 22 & 87 & 60 & 55 \\
COLING 20 & 31 & 25 & 23 \\
ACL 17 & 20 & 16 & 16 \\
CoNLL 16 & 12 & 7 & 4 \\
ESD 23 & 17 & 10 & 10 \\
ESurf 23 & 16 & 16 & 16 \\
F1000 22 & 27 & 14 & 12 \\
\midrule
Total & \textbf{579} & \textbf{414} & \textbf{383} \\ \bottomrule
\end{tabular}
\caption{Number of questions with answer evidence that could be mapped to the PDF extracted text.}
\label{tbl:dataset-numbers-with-extraxted}
\end{table}

\section{Pre- \& Post-Processing Prompts}
\subsection{Question Clean-Up \& Decontextualization}\label{sec:appendix-question-processing-clean-up-and-decontextualization}
Given the extracted question and previous sentences (\texttt{context}) from the peer review, we use the following prompt to decontextualize the question:\\
\texttt{This is part of a scientific peer review where the reviewer raises a question regarding the paper.\\
"""\\
\{context\} \{question\}\\
"""\\
Write the last question such that it can be comprehended independently without the context of the review. Resolve any references to the review. Respond with a single question.
}
\subsection{Question Decomposition}\label{sec:appendix-question-processing-decompoistion}
In case the constituency parser detects a conjunction, we use the following prompt to decompose the question:\\
\texttt{This is a sentence from a peer review containing two questions.\\
"""\\
\{question\}\\
"""\\
Write the questions such that each can be comprehended independently without the context of the other question. Resolve any references in the second question. Therefore, the fundamental question information needs to be duplicated in each question.\\
}

\subsection{Answer Free-Form Augmentation with Evidence}\label{sec:appendix-answer-augmentation-with-evidence}
To ensure a similar quality and verbosity of answers, we augment the free-form answers provided by the authors using the prompt below in case the question has annotated evidence. If it does not have annotated evidence, we use the prompt in \S\ref{sec:appendix-answer-augmentation-without-evidence}.\\
\texttt{You are a helpful scientific research assistant. Your task is to write clean answers, given noisy answers from a scientific question answering dataset. The question has been asked during a peer review of a scientific article. Given the question, background information extracted from the paper, and a noisy answer, your task is to write a clean answer. Write a concise answer that directly answers the question. Make sure all information in your answer is covered by the background. Incorporate additional information from the original answer. Write the answer neutrally, i.e., as a third person (and not the author) answering the question. For example, use "The authors" instead of "We".\\
Question: \{question\}\\
Background: \{evidence\}\\
Original Answer: \{answer\}\\
Rephrased Answer:\\
}

\subsection{Answer Free-Form Augmentation without Evidence}\label{sec:appendix-answer-augmentation-without-evidence}
\texttt{You are a helpful scientific research assistant. Your task is to write clean answers, given noisy answers from a scientific question answering dataset. The question has been asked during a peer review of a scientific article. Given the question and a noisy answer, your task is to write a clean answer. Write a concise answer that directly answers the question. Incorporate the information from the original answer. Write the answer neutrally, i.e., as a third person (and not the author) answering the question. For example, use "The authors" instead of "We".\\
Question: \{question\}\\
Original Answer: \{answer\}\\
Rephrased Answer:\\
}

\section{Question Grounding}\label{sec:cosine-similarity-question-context}
\begin{figure}[htpb]
    \centering
    \includegraphics[width=\columnwidth]{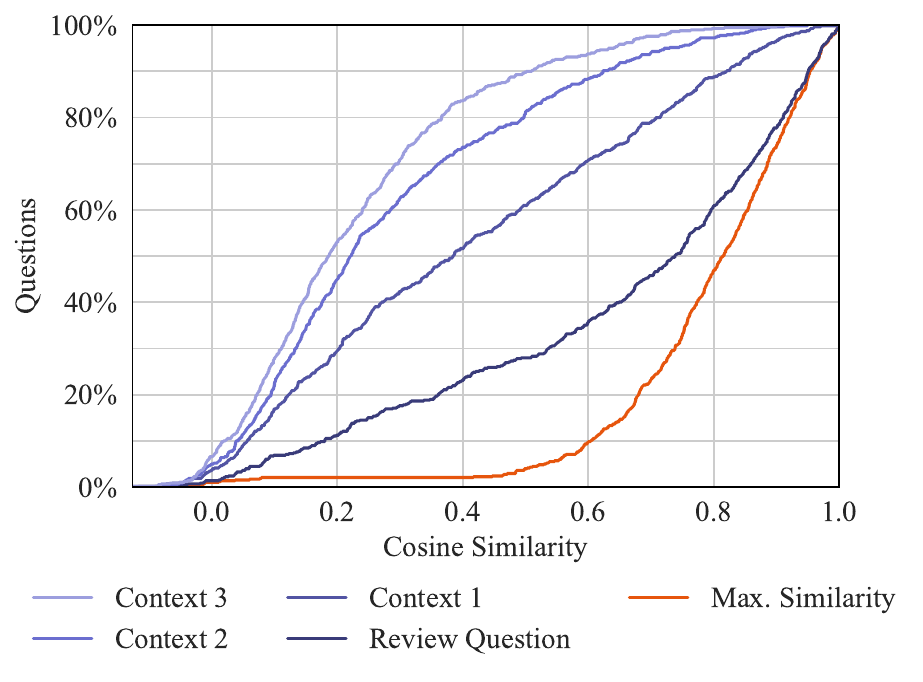}
    \caption{Empirical cumulative distribution function of the cosine similarity between the processed question and the sentences in the review. \textit{Context n} refers to the n-th preceding sentence before the raw, unprocessed \textit{Review Question}. \textit{Max. Similarity} takes the max operation over these four similarity scores, i.e., reports the similarity the processed question is most similar to.}
    \label{fig:cosine-similarity-question-context}
\end{figure}

Figure~\ref{fig:cosine-similarity-question-context} visualizes the similarity between the processed question and original review sentences. We use \texttt{all-MiniLM-L6-v2}\footnote{\url{https://huggingface.co/sentence-transformers/all-MiniLM-L6-v2}} to compute the similarity. As detailed in \S\ref{sec:data-collection}, we extract questions from the peer review and contextualize them with the preceding three sentences from the review. To understand whether our preprocessing has altered the original question or not, we compute the maximum similarity between the final processed question and the four sentences of the review (i.e., the question and the three proceeding questions). We find that $90\%$ of questions have a similarity of at least $0.60$, and $50\%$ are more than $0.82$ similar to the final processed question. This shows the quality of our cleaning, decontextualization, and decomposition steps: Questions are generally highly similar and, therefore, grounded in the original peer review.

\begin{figure*}[htpb]
    \centering
    \includegraphics[width=\textwidth]{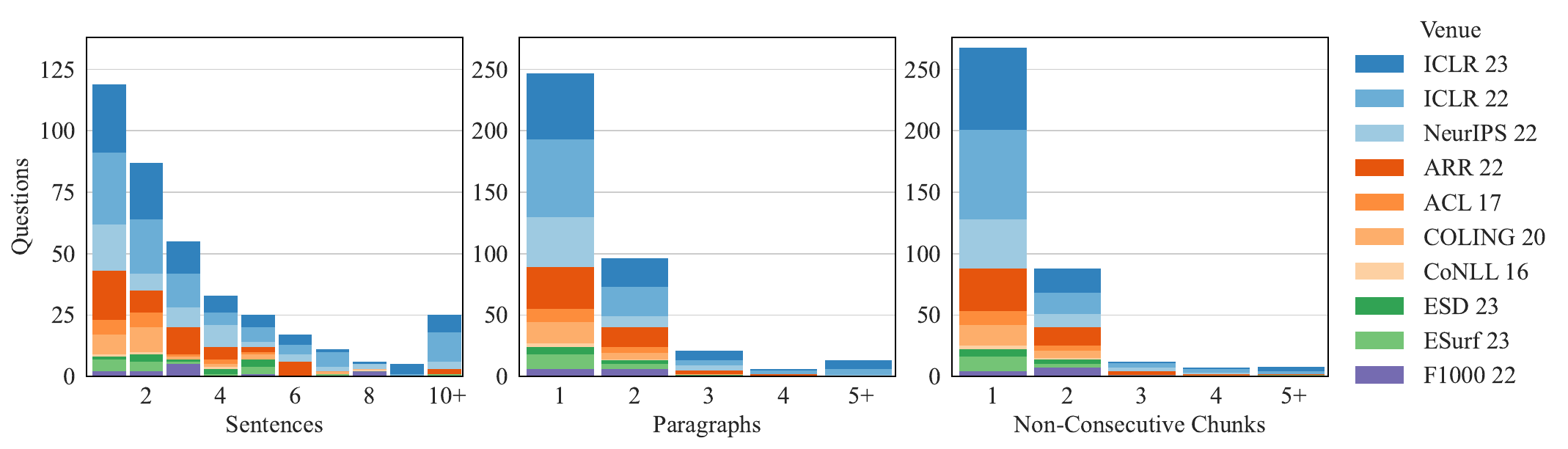}
    \caption{Number of evidence sentences (left), paragraphs (middle), and non-consecutive chunks (right) per question with annotated answer evidence.}
    \label{fig:answer-evidence-stats}
\end{figure*}

\section{Unlabeled Data}\label{sec:appendix-unlabeled-data}
\begin{table}[!h]
\small
\centering
\begin{tabular}{lrr}
\toprule
 Venue & Questions & Papers \\
\midrule
ICLR 23 & 5199 & 1188 \\
ICLR 22 & 3987 & 824 \\
NeurIPS 22 & 1186 & 110 \\
ARR 22 & 470 & 188 \\
COLING 20 & 70 & 33 \\
ACL 17 & 147 & 54 \\
CoNLL 16 & 3 & 3 \\
ESurf 23 & 312 & 51 \\
ESD 23 & 246 & 48 \\
F1000 22 & 347 & 124 \\
\midrule
\textbf{Total} & 11967 & 2623 \\
\bottomrule
\end{tabular}
\caption{Number of unlabeled questions and papers per venue.}
\label{tbl:appendix-unlabled-questions-count}
\end{table}

Besides the 579 questions with answer annotations, we additionally release all preprocessed and filtered 12k questions from 2.6k papers that have not been answered. Table~\ref{tbl:appendix-unlabled-questions-count} shows the breakdown per venue.

\section{Answer Evidence Statistics}\label{sec:appendix-answer-evidence-stats}
Figure~\ref{fig:answer-evidence-stats} reports the number of answer evidence depending on the retrieval unit. Note that this only includes answer evidence that we could map into the text extracted from the PDF. Non-consecutive chunks are essentially the number of different locations in the paper with answer evidence.

While the answer evidence for most questions comes from a single place, 30\% of questions have more than one location in the paper that addresses the question. While requiring to retrieve from multiple sources is related to multi-hop question answering \citep{welbl-etal-2018-constructing, yang-etal-2018-hotpotqa}, our setup is slightly different. We have also investigated the performance of questions with single vs multiple answer evidence chunks and have not found consistent differences. The information in the different chunks is not necessarily complementary, but it can also be that similar information is contained in each chunk, or a single chunk is sufficient to answer the question. 

\section{RAG Recall@k}
\begin{figure}[!htpb]
    \centering
    \includegraphics[width=\columnwidth]{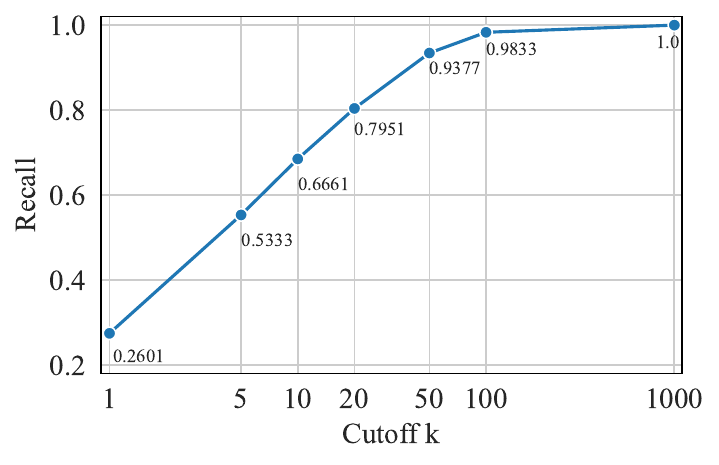}
    \caption{Recall@k of \texttt{SPLADEv3} on the answer evidence retrieval task in the paragraph setting. The paragraphs retrieved by \texttt{SPLADEv3} are used in the RAG experiments.}
    \label{fig:recall-at-k-splade-v3}
\end{figure}

Figure~\ref{fig:recall-at-k-splade-v3} shows the recall at various cutoffs $k$ for \texttt{SPLADEv3}, the best-performing model answer evidence retrieval task. This model is used as a retrieval model for the retrieval augmented answer generation experiments.

\section{Answerability Prompts}\label{sec:appendix-answerability-prompt}
We use the following prompts to determine whether a question is answerable or not in the setting where we provide the full text (\S\ref{sec:appendix-answerability-prompt-full-text}), the gold or retrieved paragraphs (\S\ref{sec:appendix-answerability-prompt-rag}).
\subsection{Full-Text}\label{sec:appendix-answerability-prompt-full-text}
\texttt{Read the following paper and answer the question. If the paper does not answer the question, answer with "No Answer".\\
Question: \{question\}\\
Paper: \{paper\}\\
Answer:
}
\subsection{RAG}\label{sec:appendix-answerability-prompt-rag}
\texttt{Read the following paragraphs of a paper and answer the question. If the paragraphs do not provide any information to answer the question, answer with "No Answer".\\
Question: \{question\}\\
Paragraphs: \{paragraphs\}\\
Answer:
}

\section{Answer Generation Prompts}\label{sec:appendix-answer-generation-prompt}
We use the following prompts to generate answers in the full-text (\S\ref{sec:appendix-answer-generation-prompt-full-text}) and RAG (\S\ref{sec:appendix-answer-generation-prompt-rag}) setting.
\subsection{Full-Text}\label{sec:appendix-answer-generation-prompt-full-text}
\texttt{Read the following paper and answer the question.\\
Question: \{question\}\\
Paper: \{paper\}\\
Answer:
}
\subsection{RAG}\label{sec:appendix-answer-generation-prompt-rag}
\texttt{Read the following paragraphs of a paper and answer the question.\\
Question: \{question\}\\
Paragraphs: \{paragraphs\}\\
Answer:
}

\section{Model Sizes and Computational Resources}
\paragraph{Answer Retrieval}
The number of parameters for each retrieval model is reported in Table~\ref{tbl:ir-models-parameters}. The retrieval experiments have been conducted on a Titan RTX 24GB.
\begin{table}[!h]
\small
\centering
\begin{tabular}{@{}lr@{}}
\toprule
Model & Parameters (M) \\ \midrule
MiniLM-L12-v2 & 33 \\
Contriever & 110 \\
Dragon+ & 110 \\
GTR-XL & 1240 \\
ColBERTv2 & 110 \\
BM25 & -- \\
SPLADEv3 & 110 \\ \bottomrule
\end{tabular}
\caption{Number of parameters in Millions of the evaluated retrieval models.}\label{tbl:ir-models-parameters}
\end{table}

\paragraph{Answerability \& Answer Generation}
Sizes for the models used in the answerability and answer generation task are reported with the model names. The number of parameters for the proprietary \texttt{GPT-3.5} and \texttt{GPT-4o} models are unknown, and we use it via the Azure API. We deploy the other models on a single A100 80GB GPU, except \texttt{Command-R} for which we require 2 A100 GPUs to fit also the longest paper fully into memory. All generation experiments use greedy decoding and use the \texttt{vllm} library \citep{kwon2023efficient} in version 0.4.2.

\section{Evaluation Metric Details}\label{sec:appendix-evaluation-metric-details}
\paragraph{Answer Evidence Retrieval}
To evaluate the answer evidence retrieval task, we use the mean reciprocal rank and recall implemented by the \texttt{pytrec\_eval} \citep{VanGysel2018pytreceval} package in version 0.5.
\paragraph{Un/Answerability} To compute the precision, recall, accuracy and F1 scores of the question answerability task, we use the classification report provided by \texttt{scikit-learn} \citep{scikit-learn} version 1.4.0.
\paragraph{Free-Form Answer Generation} The generated answers are evaluated with Rouge \citep{lin-2004-rouge}, AlignScore \citep{zha-etal-2023-alignscore} and Prometheus \citep{kim-etal-2024-prometheus}. 
For Rouge, we use the longest common subsequence (Rouge-L) between the generated answer and the reference answer. We use the \texttt{rouge-score} package in version 0.1.2 via Hugging Face's \texttt{datasets} package \cite{lhoest-etal-2021-datasets}. We also stem the generated and reference answer before computing the metric with the Porter Stemmer. All reported Rouge-L scores are F1 metrics.
For AlignScore, we use the fine-tuned checkpoint based on RoBERTa-large \citep{roberta} and use the \texttt{nli\_sp} mode, which splits the generation into sentences and uses a 3-way classification head to obtain scores. We use the original implementation in version 0.1.3.\footnote{\url{https://github.com/yuh-zha/AlignScore/commit/a0936d5afee642a46b22f6c02a163478447aa493}} For Prometheus, we use the \texttt{prometheus-eval} package with version 0.1.20\footnote{\url{https://github.com/prometheus-eval/prometheus-eval/releases/tag/v0.1.20}} and the  7B-v2.0 model\footnote{\url{https://huggingface.co/prometheus-eval/prometheus-7b-v2.0}} and instruct the model to evaluate the correctness with respect to the reference answer. Following \citet{kim-etal-2024-prometheus} and the score rubric proposed by \citet{rag-rubic-haystack} we use the following prompt:\\
\texttt{
\#\#\#Task Description:\\
An instruction (might include an Input inside it), a response to evaluate, a reference answer that gets a score of 5, and a score rubric representing a evaluation criteria are given.\\
1. Write a detailed feedback that assess the quality of the response strictly based on the given score rubric, not evaluating in general.\\
2. After writing a feedback, write a score that is an integer between 1 and 5. You should refer to the score rubric.\\
3. The output format should look as follows: "(write a feedback for criteria) [RESULT] (an integer number between 1 and 5)"\\
4. Please do not generate any other opening, closing, and explanations.\\
\#\#\#The instruction to evaluate:\\
Your task is to evaluate the generated answer against the reference answer for the question: \{question\}\\
\#\#\#Response to evaluate:\\
\{generation\}\\
\#\#\#Reference Answer (Score 5):\\
\{reference answer\}\\
\#\#\#Score Rubrics:\\
Correctness\\
Score 1: The answer is not relevant to the question and does not align with the reference answer.\\
Score 2: The answer is relevant to the question but deviates significantly from the reference answer.\\
Score 3: The answer is relevant to the question and generally aligns with the reference answer but has errors or omissions.\\
Score 4: The answer is relevant to the question and closely matches the reference answer but is less concise or clear.\\
Score 5: The answer is highly relevant, fully accurate, and matches the reference answer in both content and clarity.\\
\#\#\#Feedback:
}

\paragraph{Human Evaluation of AlignScore}
\citet{zha-etal-2023-alignscore} has evaluated correlation with human judgments extensively, particularly on factual consistency datasets. To show the reliability of AlignScore on our data, we manually label 100 randomly generated answers. Specifically, we compare the free-form and generated answer on a 1-5 Likert scale, evaluating whether the generation matches the free-form answer. This yields a significant (p<0.01) Spearman correlation of 0.449, indicating a moderate alignment between Human evaluation and the AlignScore metric.
\clearpage
\onecolumn

\section{Annotation Instructions and Interface}
\subsection{Contact Email}\label{sec:appendix-contact-email}
\begin{figure*}[ht]
    \centering
    \includegraphics[width=\textwidth]{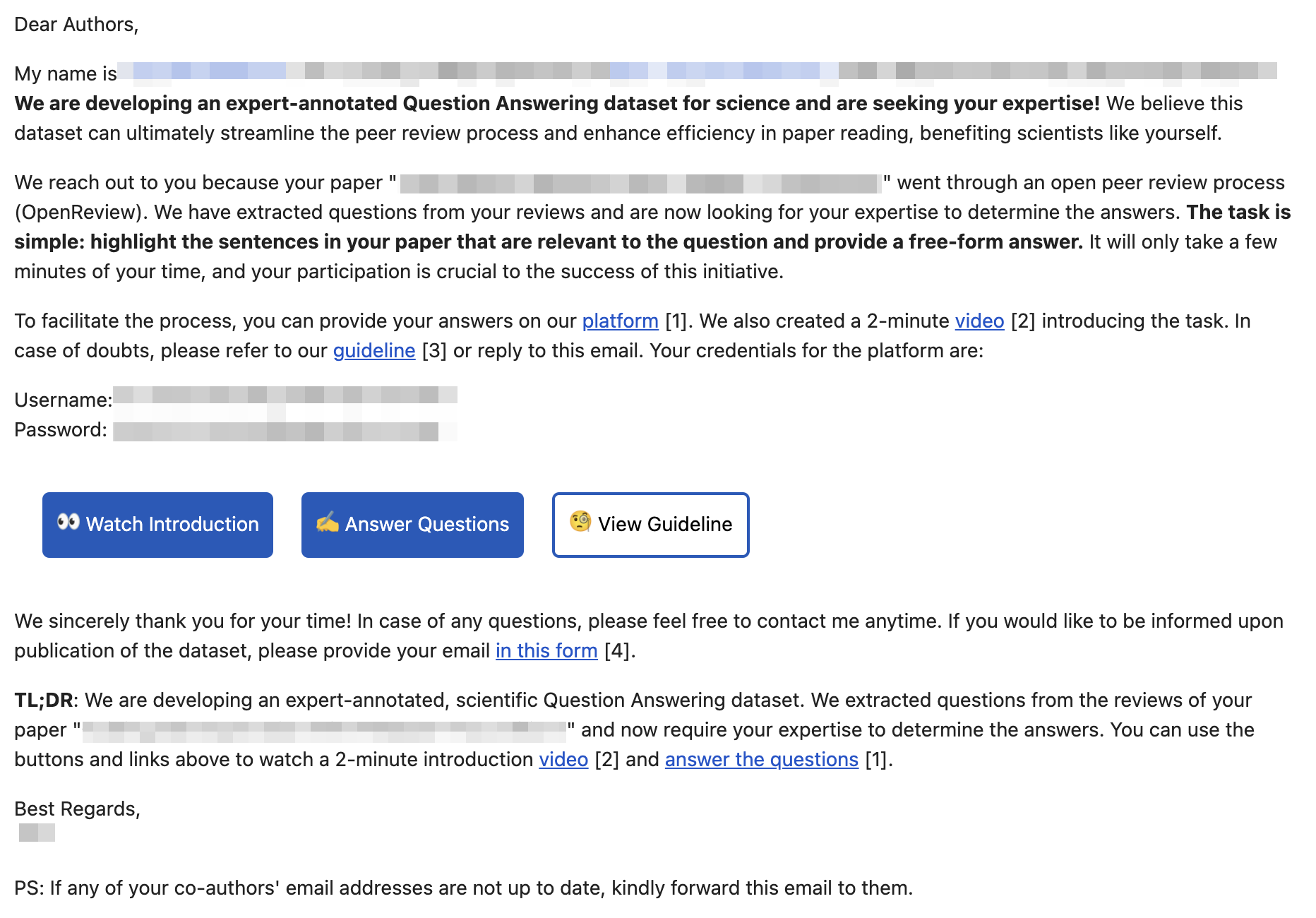}
    \caption{Exemplary contact email that has been sent to authors requesting their participation in answering the questions.}
    \label{fig:email}
\end{figure*}

\begin{multicols}{2}
Figure~\ref{fig:email} shows an instance of an email that invited the authors to participate in the data collection. Email addresses have been extracted from papers or, in the case of EGU and F1000, addresses to corresponding authors are provided online. To prevent spamming authors, we have ensured that no author received more than 3 emails (e.g., when they were listed as authors on multiple papers of a venue). Email addresses were only used to contact authors and are not part of the dataset. We also do not publicize the code to extract email addresses from papers.
\end{multicols}

\clearpage
\begin{multicols}{2}
\subsection{Annotation Interface}\label{sec:appendix-annotation-interface}
\begin{figure*}[ht]
    \centering
    \includegraphics[width=\textwidth]{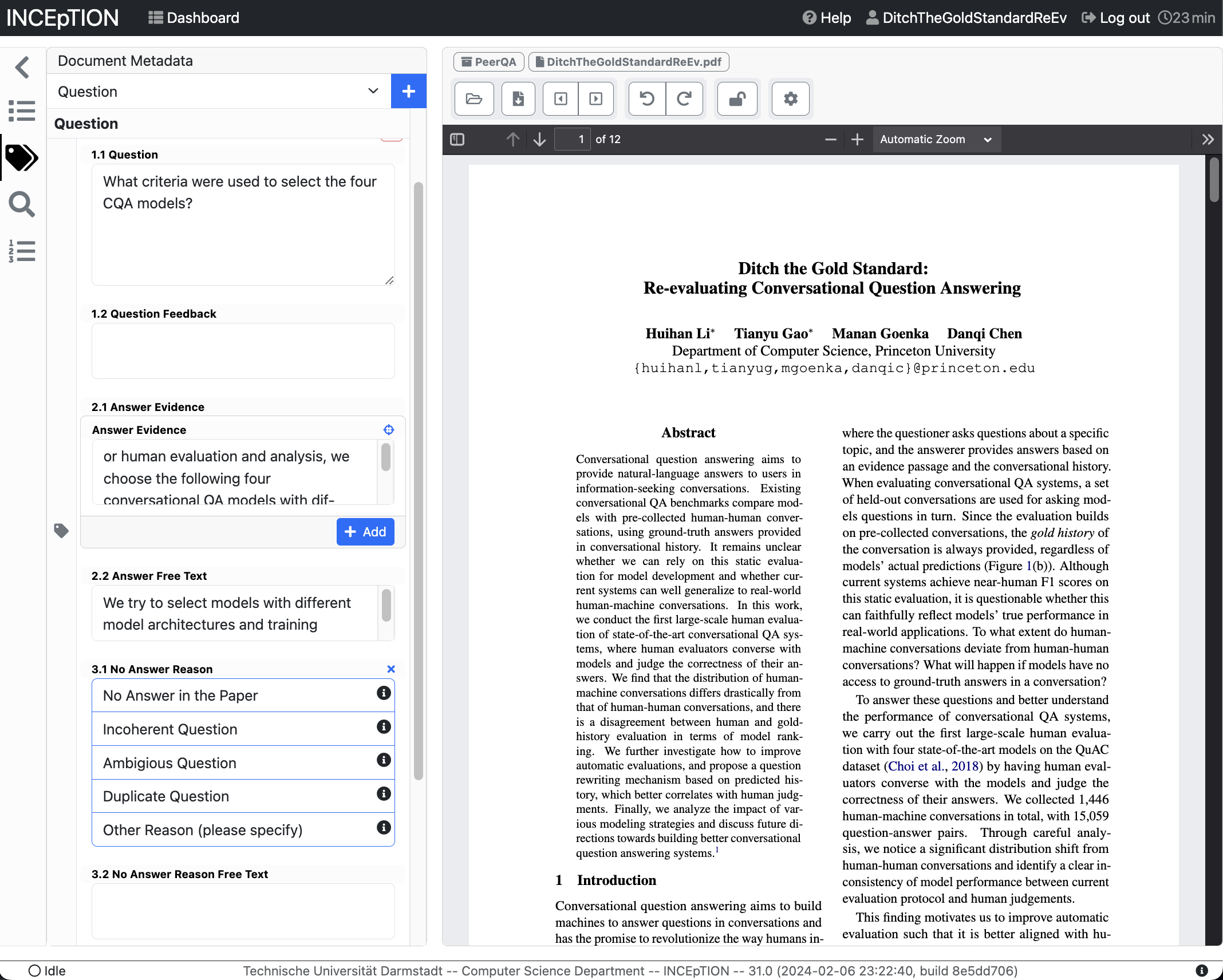}
    \caption{Screenshot of the annotation interface. The annotation consists of four parts. First, the annotator can provide feedback to the question, e.g., to correct its meaning or provide their interpretation. Second, answer evidence is annotated by highlighting sentences in the PDF. Third, a free-form answer can be provided, directly answering the question. Lastly, if a question is unanswerable or is of low quality, the interface provides an option to flag the question.}
    \label{fig:annotation-interface}
\end{figure*}

The annotation interface for providing answers is shown in Figure~\ref{fig:annotation-interface}. The camera-ready PDF of the publication is shown on the right-hand side, while answer annotations can be provided on the left side. In \textit{1.2 Question Feedback}, authors can leave free-form feedback about the question, e.g., if it should be removed or modified. By clicking on the \textit{Add} button in \textit{2.1 Answer Evidence}, text spans in the PDF can be highlighted. One highlight can span over several sentences or even pages. Multiple spans can be added by clicking the \textit{Add} button again. In \textit{2.2. Answer Free Text}, the free-form answer to the question can be given. Finally, in \textit{3.1.}, the authors can also mark the question as unanswerable or provide further feedback to the question. If none of the categories apply, feedback on why the question is unanswerable can also be provided in \textit{No Answer Reason Free Text}.
\end{multicols}

\clearpage
\begin{figure*}[!b]
    \centering
    \includegraphics[width=\textwidth]{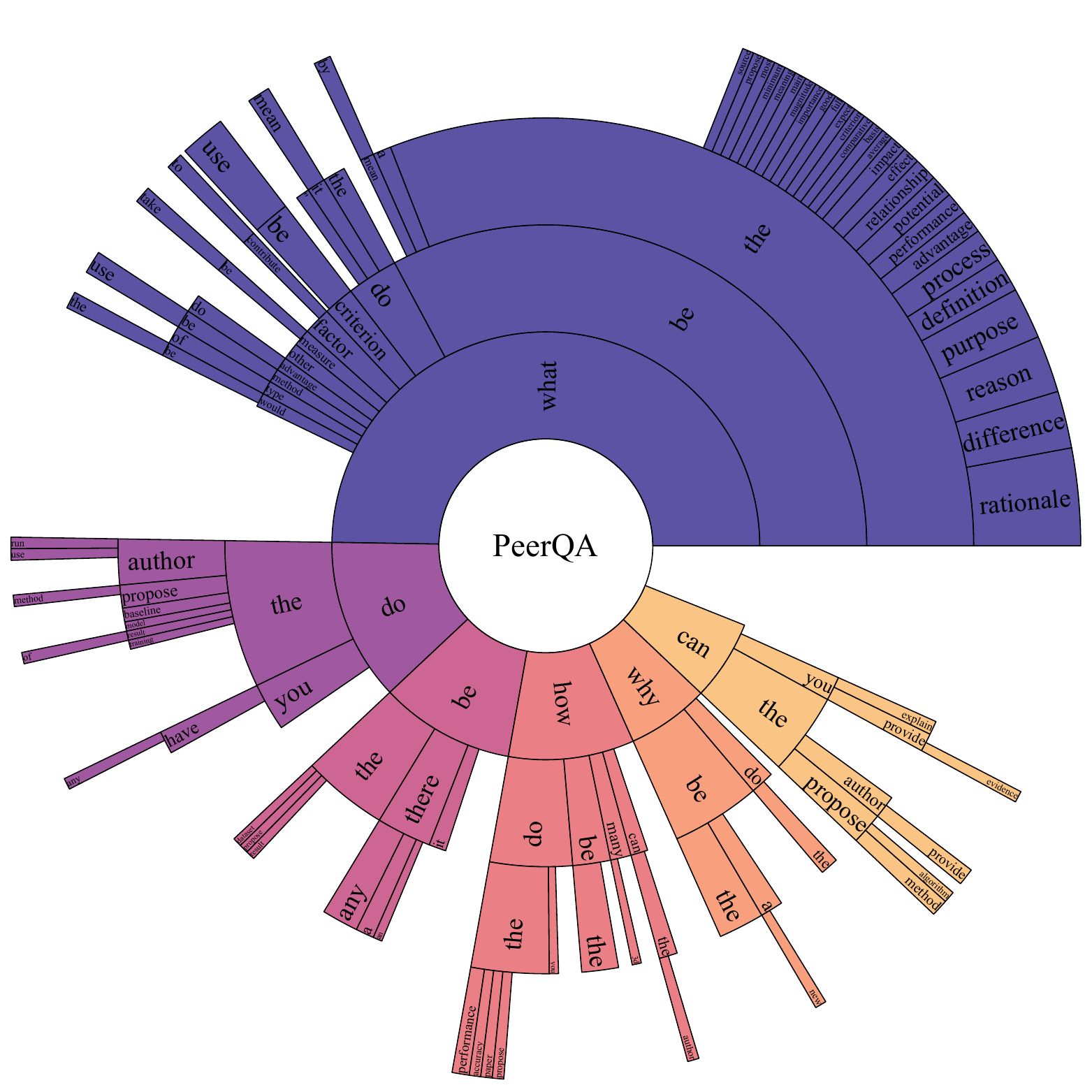}
    \caption{Sunburst diagram of the 4-grams in the PeerQA questions.}
    \vspace{11em}
    \label{fig:sunburst-questions}
\end{figure*}
\begin{multicols*}{2}
\section{Question Sunburst}\label{sec:appendix-sunburst-questions}
We visualize the starting 4-grams of all questions in Figure~\ref{fig:sunburst-questions}. All words have been lowercased and lemmatized, and rare n-grams have been discarded.
\end{multicols*}

\clearpage

\begin{figure}[!b]
     \centering
     \begin{subfigure}[b]{\textwidth}
         \centering
         \includegraphics[width=\textwidth]{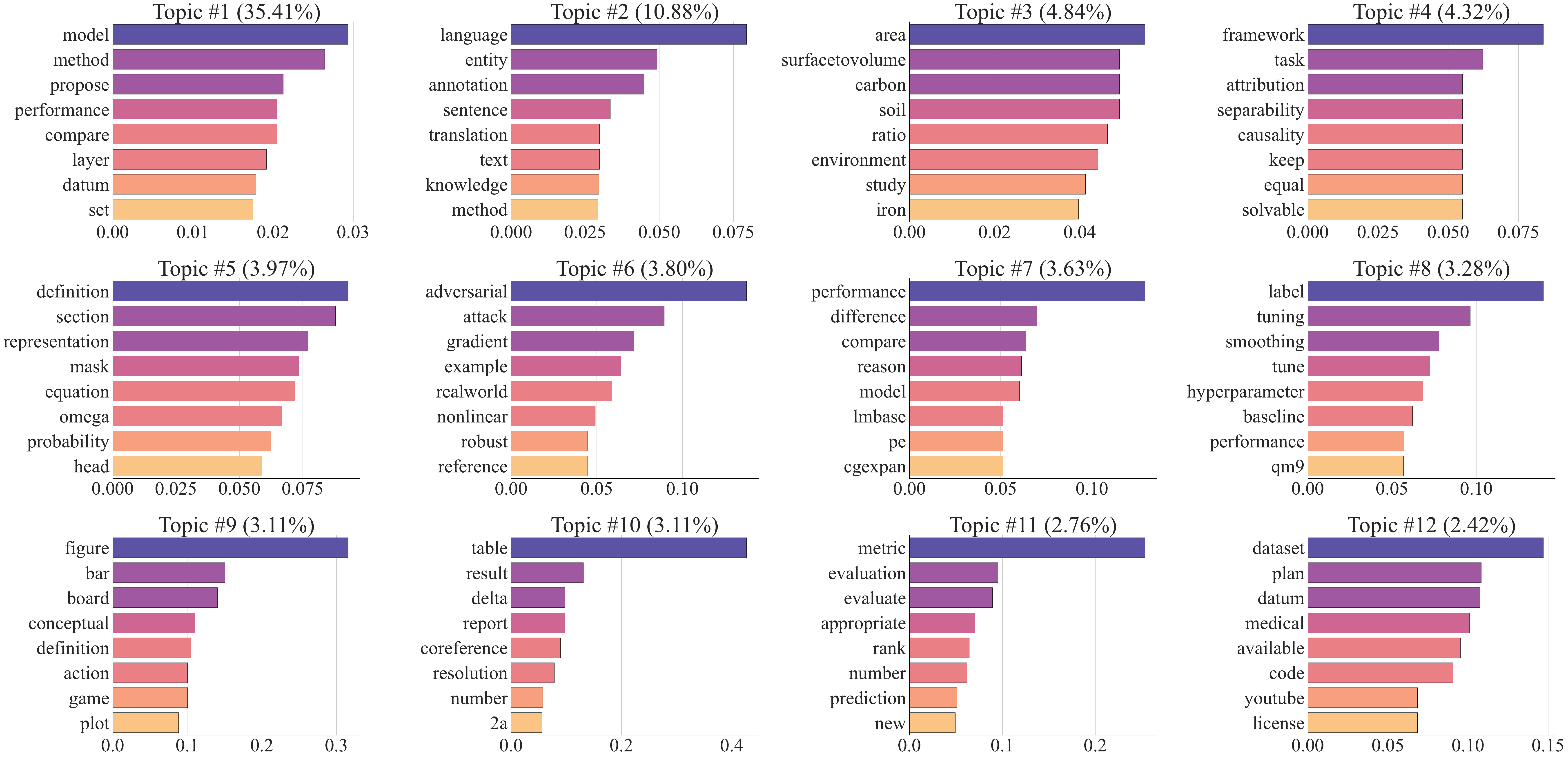}
         \caption{Topics of Labeled Questions}
         \label{fig:question-topics-labelled}
     \end{subfigure}
     \hfill
     \vspace{0.5em}
     \begin{subfigure}[b]{\textwidth}
         \centering
         \includegraphics[width=\textwidth]{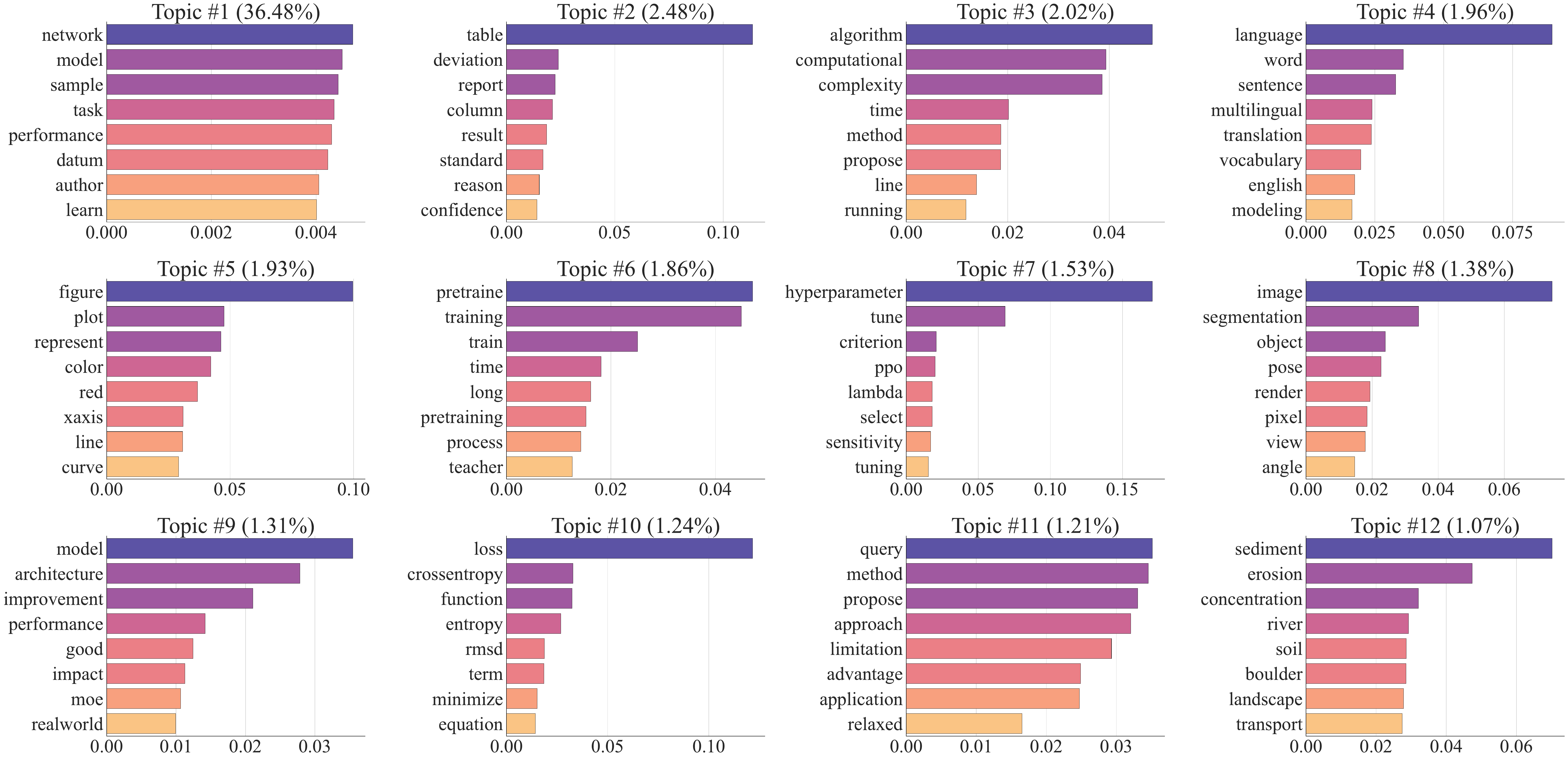}
         \caption{Topics of Unlabeled Questions}
         \label{fig:question-topics-unlabelled}
     \end{subfigure}
        \caption{Top 8 keywords for the top 12 question topics. The first topic contains all questions that could not be assigned during clustering. The bar shows the keyword's c-TF-IDF score. The top figure shows the topics for the labeled questions and the bottom for the unlabelled.}
        \label{fig:question-topic-keywords}
\end{figure}

\begin{multicols*}{2}
\section{Question Topics}\label{sec:appendix-question-topcis}
Figure~\ref{fig:question-topic-keywords} reports the result of applying BERTopic \citep{grootendorst-bertopic-2022} on the labeled (\ref{fig:question-topics-labelled}) and unlabeled (\ref{fig:question-topics-unlabelled}) questions. Table~\ref{tbl:question-topics} additionally shows representative questions for the topics. We use the standard \texttt{all-MiniLM-L6-v2} sentence transformer model to compute embeddings. After embedding, stopwords have been removed, and the words have been lemmatized using spacy \citep{spaCy2020} to improve the keyword extraction. In \S\ref{sec:analysis}, we analyzed the topics of the labeled questions cluster. We found them to be focused on the scientific community and its subtopics or related to elements in the paper. We observe similar clusters in the unlabeled data (e.g., topic \#4 for NLP, topic \#12 for Geoscience, \#6 focusing on (pre-)training, \#5 on figures and plots).
\end{multicols*}

\clearpage

\begin{table*}[!ht]
\footnotesize
\centering
\begin{tabularx}{\textwidth}{p{0.2cm}p{0.7cm}X}
\toprule
\# & Size & Representative Questions \\
\midrule
1 & 35.4\% & Why does the baseline have significantly better performance on ACE 2004 and ACE 2005 compared to Yu et al. (2020), but similar performance on OntoNotes 5 and CoNLL 2003? Does the proposed alternate way to use linear by computing the mean absolute value of the weights associated with it differ from the original linear model proposed by Dalvi et al. (2019)? Is the design of the proposed method arbitrary for all layers of a given VIT model, or are some layers fixed? \\ \midrule
2 & 10.9\% & How does your method differ from existing methods for visual and language understanding with multilinguality, such as VQA, captioning, and retrieval? What other downstream tasks, such as natural language inference, question answering, and semantic role labeling, have been tested using an encoder that has been transferred from language 1 to language 2 without any parameter updates? How can the authors ensure that the natural language sentences produced from the "ground truth" activity graphs accurately describe the scene? \\ \midrule
3 & 4.8\% & Is the authors' conclusion about the accuracy of the CMIP6 climate models in simulating the processes based on the agreement between the observed data and the models' predictions in terms of the residual variability? Do the authors assume that iron is sourced from the platform when considering the feasibility of coastal seaweeds, which have a very low surface-to-volume ratio, competing for iron against the typically small and specialized open ocean phytoplankton that have a high surface-to-volume ratio? Can we assume that coastal seaweeds, which have a very low surface-to-volume ratio, would be competitive in iron uptake against the mostly small and specialized open ocean phytoplankton that have a high surface-to-volume ratio, especially in iron-limited areas? \\ \midrule
4 & 4.3\% & Can the authors provide a justification for why only four datasets were used to evaluate the visual search models, rather than a more diverse collection of datasets? Why should associations that are solvable by AI be kept in the framework, when the purpose of the framework is to collect associations that are difficult for models but solvable by humans? Does the proposed approach address issues related to assigning different attributions to features that have the same effect on the model or assigning positive attributions to features with no effect? \\ \midrule
5 & 4.0\% & How does the paper incorporate section titles into the BOS representation? What is the purpose of multiplying the scalar sc(omega, q) by the inner product of omega and q in equation 5? What is the difference between the \verb|\odot| and \verb|\cdot| symbols in the equation for computing the overall source mask from the k masks? \\ \midrule
6 & 3.8\% & Is it possible to consistently find perturbations to empirically robust adversarial examples that result in a correctly classified image? How does the paper define the concept of an "adversarial L2 ball" when it appears to suggest that every sample should have the same classification as \verb|\tilde{x}|, contrary to the expectation that each sample within the ball should have a different classification compared to x? Could the authors provide further justification for their claim that the gradient-based attack is responsible for the shift between test and training data observed in the adversarial attack? \\ \midrule
7 & 3.6\% & What is the performance of larger GLM models compared to state-of-the-art results, given that hardware resources do not appear to be a constraint? What is the expected relationship between the performance of the algorithms and the number of updates per sample, memory size, and batch size? What could explain the difference in performance between the DICTA test set and the new test set, particularly the difference between the cha and wor scores? \\ \midrule
8 & 3.3\% & What protocol did you use to decide when to stop training and to select hyperparameters for each dataset when no labeled target data is available? Does label smoothing always improve performance, or are there cases where it can degrade performance? Does label smoothing always improve the performance of the hyperparameter-fine tuning procedure? \\ \midrule
9 & 3.1\% & What are the vertical uncertainty bars in Figure 13? What would be the correct classification for the image in Figure 1 where the space bar is hidden? What is the reason for the sudden change in the green and blue curves in Figure 2 at epoch 90? \\ \midrule
10 & 3.1\% & What is the impact of adjusting $\delta$ on the results of Table 1? Is the coreference resolution pipeline depicted in Table 1 universally accepted in the field of coreference resolution? What is the difference between the results in Table 1 and Table 2(a)? \\ \midrule
11 & 2.8\% & Is there an optimal number of MSD points to use in order to minimize the error on the estimated parameters, and is there an option to automatically determine this number? Why is a new metric, concept purity, introduced instead of using the same set of metrics provided in Yuan et al. (2020)? What benefits does improving the upper bound for the Information Gain evaluation metric provide in practice? \\
\midrule
12 & 2.4\% & Are all participants in the trial pregnant women who are less than 36 weeks gestation? What is the rationale for changing the data distribution if the KB was compiled by medical papers? What ethical considerations were taken into account when selecting the data for the dataset? \\ 
\bottomrule
\end{tabularx}
\caption{Questions clustered into the top 12 topics by BERTopic. \textit{Representative Questions} are automatically identified by BERTopic.}
\label{tbl:question-topics}
\end{table*}

\clearpage

\section{Example Annotations}\label{sec:appendix-examplary-annotation}
\begin{table*}[!h]
\scalebox{0.98}{
\begin{minipage}{\linewidth}
\centering
\begin{adjustbox}{angle=270} 
\tiny
\begin{tabularx}{0.93\textheight}{@{}p{0.1cm}p{3.1cm}Xp{3.25cm}p{6cm}@{}}
\toprule
 & Question & Answer Evidence & Free-Form Answer & GPT-4 Augmented Answer \\ \midrule
\multirow{1}{*}{\rotatebox{90}{ICLR 23}} 
& What benefits does improving the upper bound for the Information Gain evaluation metric provide in practice? & to measure the quality of such models An interpretable alternative evaluation metric for likelihood-based models is Normalized Information Gain ( NInGa ) & interpretable and robust evaluation of full likelihood models & Improving the upper bound for the Information Gain evaluation metric provides the benefit of enabling a more interpretable and robust evaluation of full likelihood models in practice. \\ \midrule
\multirow{1}{*}{\rotatebox{90}{ICLR 22}} & Does the expected label always match the most probable label given the noisy data? & This makes it clear that, overall, the pattern zig-zags, first moving towards the unknown true label before eventually turning to memorize the wrong label & No, that is why zig-zag might happen for the hard labels. & No, the expected label does not always match the most probable label given the noisy data. This is indicated by the observed zig-zag pattern, where the process initially moves towards the true label but eventually turns to memorize the incorrect label. \\ \midrule
\multirow{1}{*}{\rotatebox{90}{NeurIPS 22}} & What criteria were used to select the videos and classes for the study? & This is inspired by the fact that complex human activities in real-world settings are usually hierarchical and compositional across space and time. & We selected videos and classes that contain complex human activities that are hierarchical and compositional. & The criteria for selecting the videos and classes for the study were based on the presence of complex human activities that are hierarchical and compositional in nature.\par \\ \midrule
\multirow{1}{*}{\rotatebox{90}{ARR 22}} & What impact does the layer norm approximation have on the performance of the tasks discussed in the paper? & Another challenge of THE-X is the attention score input of layer normalization.\par In most cases, the scale of multi-head attention output is very dense around [-1,1].\par However, before normalization, we also observe the attention scores are scarily sparse, with some extreme values reaching 1e4, which is difficult for our LN-distill stage.\par To prevent the overflow attention scores, we use the weight decay of Adam optimizer as regularization.\looseness=-1 & Layernorm approximation may lead to attention overflow, which dramastically affect the performance of downstream tasks & The approximation of layer normalization can potentially lead to attention overflow. This can significantly impact the performance of the downstream tasks discussed in the paper. \\ \midrule
\multirow{1}{*}{\rotatebox[origin=c]{90}{COLING 20}} & What measures were taken to ensure that paths were not shared between the training and test sets? & To avoid giving an unfair advantage to the path-based models, we filtered both validation and test sets to only include hyponym queries that are unseen anywhere in the full taxonomy paths of the training data.\par We also exclude hyponyms from the test and validation sets which appear as hyponyms in the training set7 to prevent the models from merely copying. & All queries observed during path training and all hyponyms in the training set that also appear as hyponyms in the test set are eliminated. & The authors implemented measures to ensure that paths were not shared between the training and test sets by filtering both validation and test sets to only include hyponym queries that were not present in the full taxonomy paths of the training data. Additionally, any hyponyms that appeared in the training set were excluded from the test and validation sets to prevent the models from simply replicating the data. \\ \midrule
\multirow{1}{*}{\rotatebox{90}{ACL 17}} & What would be the impact of not having access to pre-trained embeddings on the performance of this method? & Our main contribution is to show that the context sensitive representation captured in the LM embeddings is useful in the supervised sequence tagging setting. When we include the LM embeddings in our system overall performance increases from 90.87\% to 91.93\% F 1 for the CoNLL 2003 NER task, a more then 1\% absolute F1 increase, and a substantial improvement over the previous state of the art.  We also establish a new state of the art result (96.37\% F 1 ) for the CoNLL 2000 Chunking task. Importantly, the LM embeddings amounts to an average absolute improvement of 1.06 and 1.37 F 1 in the NER and Chunking tasks, respectively. & Performance decreases by about 1\% on the CoNLL 2003 NER task, and by 1.4\% on the CoNLL 2000 Chunking task when removing pretrained language model embeddings. & The absence of pre-trained language model embeddings would result in a decrease in performance by approximately 1\% on the CoNLL 2003 NER task, and by around 1.4\% on the CoNLL 2000 Chunking task. \\ \midrule
\multirow{1}{*}{\rotatebox{90}{CoNLL 16}} & Why were post-editing rates chosen over prediction (h)ter for intrinsic uncertainty evaluation? & Our decision to focus on post-editing time was based on the fact that time is a more complete measure of post-editing effort, capturing not only technical effort like HTER, but also cognitive effort ( Koponen et al., 2012 ).\par Additionally, time is more directly applicable in real translation environments – where uncertainty estimates could be useful, as it relates directly to productivity measures. & The authors state that normalised post-editing times (i.e. post-editing rates) are a more realistic measure with respect to operational effort, and can be adapted to different application scenarios. & The authors opted for post-editing rates over prediction (h)ter for intrinsic uncertainty evaluation because they believed that time is a more comprehensive measure of post-editing effort. This measure not only captures technical effort, like HTER, but also cognitive effort. Furthermore, time is more directly applicable in real translation environments, where uncertainty estimates could be beneficial, as it correlates directly to productivity measures. Thus, post-editing rates provide a more realistic measure of operational effort and can be adapted to various application scenarios.\looseness=-1 \\ \midrule
\multirow{1}{*}{\rotatebox{90}{ESD 23}} & What is the impact on detection or signal-to-noise ratio (SNR) if RR-based fingerprint construction is not applied? & robust detection and attribution (D\&A) of forced change in the water cycle are impaired by low signal-to-noise ratios.\par The concept of detection and attribution is to use climate model simulations in which the applied forcings are known and internal variability can be reduced by averaging multiple realisations, to estimate a so-called fingerprint that represents the effect of the applied forcings on climate variables of interest. & detection won't be possible if no SNR-maximising method/process is applied to the noise observational data & If the RR-based fingerprint construction is not applied, it could potentially impair the detection process due to the low signal-to-noise ratios. This could make detection impossible without the application of a method or process that maximizes the signal-to-noise ratio in the observational data.\\ \midrule
\multirow{1}{*}{\rotatebox{90}{ESurf 23}} & Does the orientation of the grid, with potentially more north-facing slopes in the Elwha area than in the Quinault area, affect the absorption of solar radiation, thereby influencing glacier distribution? & Our flowline model captures the variability in the direction of the steepest precipitation gradients but neglects variability across the valleys that may be important, especially for spatially variable drivers of ablation, such as topographic shading. & We do not include topographic shading as a factor influencing melting in the model - so the answer to this question is no - not in our model.  In the real world, the north-facing slopes of the Elwha headwaters may have been important in the mass balance of past glaciers. & The authors' model does not account for topographic shading as a factor influencing melting, thus the orientation of the grid does not affect the absorption of solar radiation in their model. However, in reality, the north-facing slopes of the Elwha headwaters may have played a significant role in the mass balance of past glaciers. \\ \midrule
\multirow{1}{*}{\rotatebox{90}{F1000 22}} & How was the conceptual model in Figure 2 developed? & 
Participants were recruited from their schools and colleges through the activity recruitment poster and “ Word-of-Mouth ” from the teachers and lecturers.\par 
Then, their crafts will be evaluated by art teachers and ranked on a leaderboard.\par The participants who were ranked on the leaderboard will receive certificates and prizes.\par After experiencing two-level EGL, 29 students were purposively selected as FGD participants.\par
Figure 2 shows the overview of the EGL activity flow. & It is based on the procedures of the EGL activity from participation to focus group discussion. & The conceptual model depicted in Figure 2 was developed based on the procedures of the Experiential Group Learning (EGL) activity. This process ranged from participant recruitment to the focus group discussion.\\ 
\bottomrule
\end{tabularx}
\end{adjustbox}
\caption{Exemplary questions, answer evidence, and free-form answers of the PeerQA dataset from all venues.}\label{tbl:examplary-annotation}
\end{minipage}
}
\end{table*}

\clearpage
\begin{multicols}{2}
\section{Question Classes}\label{sec:appendix-question-classes}
Table~\ref{tbl:question-classes} reports the number of questions per class and representative questions for that class. The definitions for each class are the following:
\paragraph{Method Clarification} Questions to better understand a specific detail (e.g., a parameter) or inner workings of a proposed or used method, including methods used for obtaining data or details about the experiment setup/process.
\paragraph{Data Clarification} Questions to understand the process of obtaining data or properties of the used data for an experiment, however, excluding questions about a method to obtain data.
\paragraph{Justification/Rationale} Questions that challenge an assumption, ask the authors to motivate a decision's reasoning or are critical towards a process/finding.
\paragraph{Analysis} Questions asking for a better understanding of a result, e.g., why a method works or questions asking about what factors contribute to a result/finding. 
\paragraph{Implication} Questions about potential real-world applications, transfers of the data/method/findings to other applications/domains/tasks, or wider-scoped consequences of the findings.
\paragraph{Definition} Questions about the (intended) meaning of a certain phrase or term used in the paper.
\paragraph{Comparison} Questions asking for comparisons or differences between methods/data or different studies.
\paragraph{Evaluation/Evidence} Questions asking for details about a result (excluding analysis of results), details of the evaluation process, or evidence to support a certain claim.
\end{multicols}

\begin{table*}[ht]
\small
\centering
\begin{tabularx}{\textwidth}{@{}p{2.5cm}p{1cm}X@{}}
\toprule
Class & Size & Representative Questions \\ \midrule
Method Clarification & 31\% & How was the fine tuning done for the step sizes in the experiments?, Did the baselines in both experiments 1 and 2 only use a single seed? What is the set of signed input gradients in the second paragraph of section 4.2? \\ \midrule
Data Clarification & 13\% & Do the experts who annotated the dataset have expertise in linguistics or in the domain of the dataset? What is the time resolution of the forcing data used in the study, specifically, is it daily? Do the vocabulary items of the templates used in the paper have adequate representation in the training data? \\ \midrule
Justification/Rationale & 12\% & What motivated the authors to theoretically analyze the dense case and then empirically evaluate the sparse case? Are ten locations sufficient to represent the variety of surfaces in urban environments? Why is the chosen metric appropriate for evaluating the results? \\ \midrule
Comparison & 11.5\% & How does the proposed method compare to other types of vision transformers, such as Swin Transformer or Multiscale Vision Transformers? What is the difference between the MOMA dataset and the MOMA-LRG dataset? How does the performance of the filter-kd model compare to models trained using label smoothing and knowledge distillation with the optimum temperature? \\ \midrule
Analysis & 9\% & What factors influence the degree of separability when adapting a model to a task? Is it clear what the source of the improvements of Histruct+ (Roberta-base) over Bertsumext are? What factors were responsible for the success of the path-based model? \\ \midrule
Implications & 8\% & What are the potential applications of the data presented in this paper? Can the proposed data augmentation be applied to other tasks besides ILA? Do you think that the same framework on variance of ensembles would work equally well in the semantic feature space as in the space of logits? \\ \midrule
Evaluation/Evidence & 8\% & What is the evidence that the generative model is successful in synthesizing new molecules? Do you evaluate playing strength of agents by restricting them by MCTS iteration counts or by time limits? Did the authors run multiple trials to evaluate the performance of the graph-based neural network? \\ \midrule
Definition & 7.5\% & What is the definition of difficulty used in the paper to analyze the learning path of the network's predicted distribution? What is the variational approximation of c given by the query and support sets? What is the definition of $f_{i+1}$? \\ 
\bottomrule
\end{tabularx}
\caption{Distribution of question classes based on 100 questions randomly sampled from PeerQA. \textit{Representative Questions} shows manually picked questions that best correspond to the definition of the class.}
\label{tbl:question-classes}
\end{table*}

\clearpage

\begin{multicols}{2}
\section{Answerability Evaluation}\label{sec:answerability-evaluation}
Table~\ref{tbl:answerability-evaluation} shows detailed evaluation metrics for the answerability task, and Figure~\ref{fig:results-un-answerable} visualizes them. We report Precision, Recall, and F1-Score on both the answerable and unanswerable questions, as well as the average accuracy, weighted, and macro F1-Score.
\vfill\null
\columnbreak
\vfill\null
\end{multicols}
\begin{table*}[!h]
\small
\centering
\begin{tabular}{@{}lcccccccccc@{}}
\toprule
 &  & \multicolumn{3}{c}{Answerable ($N=383$)} & \multicolumn{3}{c}{Unanswerable ($N=112$)} & \multicolumn{3}{c}{Average} \\
 \cmidrule(lr){3-5}
 \cmidrule(lr){6-8}
 \cmidrule(lr){9-11}
Model & Ctx. & Prec. & Recall & F1 & Prec. & Recall & F1 & Acc. & W-F1 & M-F1 \\ \midrule
\multirow{6}{*}{\begin{tabular}[c]{@{}l@{}}Llama-3\\ IT-8B-8k\end{tabular}} 
 & G    & \textbf{1.0000} & 0.4517 & 0.6223 & --     & --     & --     & 0.4517 & 0.6223 & 0.3112 \\
 & 10   & 0.8407 & 0.3995 & 0.5416 & 0.2652 & \textbf{0.7411} & 0.3906 & 0.4768 & 0.5074 & 0.4661 \\
 & 20   & \textbf{0.8796} & 0.2480 & 0.3870 & 0.2558 & \textbf{0.8839} & 0.3968 & 0.3919 & 0.3892 & 0.3919 \\
 & 50   & 0.7907 & 0.1775 & 0.2900 & 0.2298 & \textbf{0.8393} & 0.3608 & 0.3273 & 0.3060 & 0.3254 \\
 & 100  & 0.7667 & 0.1802 & 0.2918 & 0.2247 & \textbf{0.8125} & 0.3520 & 0.3232 & 0.3054 & 0.3219 \\
 & FT   & 0.8168 & 0.5587 & 0.6636 & 0.2747 & 0.5714 & 0.3710 & 0.5616 & 0.5974 & 0.5173 \\
 \midrule
\multirow{6}{*}{\begin{tabular}[c]{@{}l@{}}Llama-3\\IT-8B-32k\end{tabular}} 
 & G & \textbf{1.0000} & 0.4047 & 0.5762 & -- & -- & -- & 0.4047 & 0.5762 & 0.2881 \\
 & 10 & 0.8326 & 0.4804 & 0.6093 & 0.2737 & 0.6696 & 0.3886 & 0.5232 & 0.5593 & 0.4989 \\
 & 20 & 0.8182 & 0.4700 & 0.5970 & 0.2618 & 0.6429 & 0.3721 & 0.5091 & 0.5461 & 0.4846 \\
 & 50 & 0.8056 & 0.3786 & 0.5151 & 0.2444 & 0.6875 & 0.3607 & 0.4485 & 0.4802 & 0.4379 \\
 & 100 & 0.7984 & 0.2585 & 0.3905 & 0.2345 & 0.7768 & 0.3602 & 0.3758 & 0.3837 & 0.3754 \\
 & FT & 0.8488 & 0.3812 & 0.5261 & 0.2663 & 0.7679 & 0.3954 & 0.4687 & 0.4965 & 0.4608 \\
  \midrule
\multirow{6}{*}{\begin{tabular}[c]{@{}l@{}}Mistral\\ IT-v02-7B-32k\end{tabular}} 
 & G    & \textbf{1.0000} & \textbf{0.8877} & \textbf{0.9405} & --     & --     & --     & \textbf{0.8877} & \textbf{0.9405} & \textbf{0.4703} \\
 & 10   & 0.7854 & \textbf{0.9269} & \textbf{0.8503} & \textbf{0.3488} & 0.1339 & 0.1935 & \textbf{0.7475} & \textbf{0.7017} & 0.5219 \\
 & 20   & 0.7790 & \textbf{0.9295} & \textbf{0.8476} & 0.2895 & 0.0982 & 0.1467 & \textbf{0.7414} & 0.6890 & 0.4971 \\
 & 50   & 0.7768 & \textbf{0.9086} & \textbf{0.8375} & 0.2553 & 0.1071 & 0.1509 & \textbf{0.7273} & 0.6822 & 0.4942 \\
 & 100  & 0.7824 & \textbf{0.9295} & \textbf{0.8496} & \textbf{0.3250} & 0.1161 & 0.1711 & \textbf{0.7455} & \textbf{0.6961} & 0.5103 \\
 & FT   & 0.7803 & \textbf{0.9739} & \textbf{0.8664} & \textbf{0.4118} & 0.0625 & 0.1085 & \textbf{0.7677} & \textbf{0.6949} & 0.4875 \\
  \midrule
\multirow{6}{*}{\begin{tabular}[c]{@{}l@{}}Command-R\\ v01-34B-128k\end{tabular}} 
 & G    & \textbf{1.0000} & 0.7232 & 0.8394 & -- & -- & -- & 0.7232 & 0.8394 & 0.4197 \\
 & 10   & 0.7985 & 0.8172 & 0.8077 & 0.3204 & 0.2946 & 0.3070 & 0.6990 & 0.6944 & \textbf{0.5574} \\
 & 20   & 0.8025 & 0.8381 & 0.8199 & \textbf{0.3474} & 0.2946 & 0.3188 & 0.7152 & \textbf{0.7065} & \textbf{0.5694} \\
 & 50   & 0.8031 & 0.8198 & 0.8114 & \textbf{0.3365} & 0.3125 & 0.3241 & 0.7051 & \textbf{0.7011} & 0.5677 \\
 & 100  & 0.7949 & 0.8094 & 0.8021 & 0.3048 & 0.2857 & 0.2949 & 0.6909 & 0.6873 & 0.5485 \\
 & FT   & 0.8113 & 0.7520 & 0.7805 & 0.3214 & 0.4018 & 0.3571 & 0.6727 & 0.6847 & \textbf{0.5688} \\
 \midrule
\multirow{6}{*}{\begin{tabular}[c]{@{}l@{}}GPT-3.5\\ Turbo-0613-16k\end{tabular}} 

 & G    & \textbf{1.0000} & 0.4935 & 0.6608 & --     & --     & --     & 0.4935 & 0.6608 & 0.3304 \\
 & 10   & 0.8107 & 0.4360 & 0.5671 & 0.2526 & 0.6518 & 0.3641 & 0.4848 & 0.5211 & 0.4656 \\
 & 20   & 0.8248 & 0.5039 & 0.6256 & 0.2720 & 0.6339 & 0.3807 & 0.5333 & 0.5702 & 0.5032 \\
 & 50   & 0.8168 & 0.5587 & 0.6636 & 0.2747 & 0.5714 & 0.3710 & 0.5616 & 0.5974 & 0.5173 \\
 & 100  & 0.8507 & 0.4465 & 0.5856 & 0.2789 & 0.7321 & 0.4039 & 0.5111 & 0.5445 & 0.4948 \\
 & FT   & 0.8348 & 0.2507 & 0.3855 & 0.2447 & \textbf{0.8304} & 0.3780 & 0.3818 & 0.3838 & 0.3818 \\
\midrule
\multirow{6}{*}{\begin{tabular}[c]{@{}l@{}}GPT-4o\\0806-128k\end{tabular}} 
 & G    & \textbf{1.0000} & 0.4465 & 0.6173 & --     & --     & --     & 0.4465 & 0.6173 & 0.3087 \\
 & 10   & \textbf{0.8439} & 0.5222 & 0.6452 & 0.2907 & 0.6696 & \textbf{0.4054} & 0.5556 & 0.5909 & 0.5253 \\
 & 20   & 0.8560 & 0.5744 & 0.6875 & 0.3151 & 0.6696 & \textbf{0.4286} & 0.5960 & 0.6289 & 0.5580 \\
 & 50   & \textbf{0.8604} & 0.5953 & 0.7037 & 0.3261 & 0.6696 & \textbf{0.4386} & 0.6121 & 0.6437 & \textbf{0.5712} \\
 & 100  & \textbf{0.8543} & 0.5666 & 0.6813 & 0.3112 & 0.6696 & \textbf{0.4249} & 0.5899 & 0.6233 & \textbf{0.5531} \\
 & FT   & \textbf{0.8458} & 0.5300 & 0.6517 & 0.2941 & 0.6696 & \textbf{0.4087} & 0.5616 & 0.5967 & 0.5302 \\

\bottomrule
\end{tabular}
\captionof{table}{Evaluation results on the answerability task of various LLMs, with different context settings (G = Gold Evidence, FT = Full-Text, 10/20/50/100 = Top-k passages). Note that the class distribution is imbalanced. There are a total of 383 answerable and 112 unanswerable questions. W-F1 is Weighted F1, M-F1 is Macro F1.}\label{tbl:answerability-evaluation}
\end{table*}

\clearpage

\begin{multicols}{2}
\section{Answer Generation Evaluation}\label{sec:answer-generation-evaluation}
Table~\ref{tbl:answer-generation-evaluation} reports the exact numbers of the free-form answer generation experiment for all models and contexts, corresponding to Figure~\ref{fig:answer-generation-eval}.
\vfill\null
\columnbreak
\vfill\null
\end{multicols}

\begin{table}[!hp]
\begin{center}
\small

\begin{tabular}{@{}lccccccccc@{}}
\toprule
 &  & \multicolumn{3}{c}{Rouge-L} & \multicolumn{3}{c}{AlignScore} & \multicolumn{2}{c}{Prometheus} \\ 
\cmidrule(lr){3-5}
\cmidrule(lr){6-8}
\cmidrule(lr){9-10}
Model & Ctx. & AE & FF & GPT-4 FF & AE & FF & GPT-4 FF & FF & GPT-4 FF \\
\midrule

\multirow{6}{*}{\begin{tabular}[c]{@{}l@{}}Llama-3\\ IT-8B-8k\end{tabular}} 
 & G    & 0.1683 & 0.2295 & 0.2569 & 0.5731 & 0.1098 & 0.2643 & 3.1102 & 3.1593 \\
 & 10   & 0.1670 & 0.2113 & \textbf{0.2479} & 0.3839 & 0.1107 & 0.2107 & 3.1347 & 3.1828 \\
 & 20   & 0.1771 & 0.2074 & 0.2458 & 0.3719 & 0.1041 & 0.1965 & 3.1878 & 3.2454 \\
 & 50   & 0.1621 & 0.2050 & 0.2357 & 0.3402 & 0.1062 & 0.1958 & 3.0122 & 3.0313 \\
 & 100  & 0.1418 & 0.2069 & 0.2278 & 0.3255 & 0.1067 & 0.2184 & 2.8082 & 2.7885 \\
 & FT   & 0.1484 & 0.1736 & 0.2037 & 0.2719 & 0.0653 & 0.1159 & 2.7510 & 2.9321 \\

\midrule

\multirow{6}{*}{\begin{tabular}[c]{@{}l@{}}Llama-3\\ IT-8B-32k\end{tabular}}
& G   & 0.1648 & 0.2286 & 0.2567 & 0.5778 & 0.1016 & 0.2436 & 3.1673 & 3.1749 \\
& 10  & 0.1513 & \textbf{0.2258} & 0.2464 & 0.3970 & 0.1142 & 0.2177 & 3.1388 & 3.1410 \\
& 20  & 0.1558 & 0.2204 & 0.2425 & 0.4001 & 0.1115 & 0.2109 & 3.1388 & 3.1227 \\
& 50  & 0.1546 & 0.2061 & 0.2397 & 0.3750 & 0.0999 & 0.2011 & 3.0571 & 3.1358 \\
& 100 & 0.1664 & 0.2099 & 0.2412 & 0.3785 & 0.1037 & 0.2008 & 3.0000 & 3.2010 \\
& FT  & 0.1835 & 0.1948 & 0.2260 & 0.3311 & 0.0711 & 0.1450 & 3.1959 & 3.2167 \\
 \midrule
\multirow{6}{*}{\begin{tabular}[c]{@{}l@{}}Mistral\\ v02-7B-32k\end{tabular}}
 & G    & \textbf{0.2442} & 0.1922 & 0.2432 & 0.6407 & 0.0827 & 0.1977 & 3.4245 & \textbf{3.4517} \\
 & 10   & \textbf{0.1967} & 0.1667 & 0.2032 & 0.3573 & 0.0612 & 0.1094 & 3.2490 & 3.3629 \\
 & 20   & \textbf{0.2039} & 0.1670 & 0.2011 & 0.3449 & 0.0505 & 0.1107 & 3.2408 & 3.2663 \\
 & 50   & \textbf{0.2023} & 0.1572 & 0.1943 & 0.3211 & 0.0496 & 0.1017 & 3.1306 & 3.1958 \\
 & 100  & \textbf{0.2023} & 0.1593 & 0.1927 & 0.3142 & 0.0634 & 0.1209 & 3.0245 & 3.0809 \\
 & FT   & \textbf{0.1883} & 0.1344 & 0.1678 & 0.2599 & 0.0328 & 0.0750 & 2.9796 & 3.1227 \\
 \midrule
\multirow{6}{*}{\begin{tabular}[c]{@{}l@{}}Command-R\\ v01-34B-128k\end{tabular}}
 & G    & 0.1310 & 0.2294 & 0.2081 & 0.5604 & 0.1362 & 0.3059 & 3.0571 & 3.0052 \\
 & 10   & 0.1211 & 0.2104 & 0.1973 & 0.3767 & 0.1221 & 0.2275 & 3.1551 & 3.1723 \\
 & 20   & 0.1220 & 0.2164 & 0.1978 & 0.3823 & 0.1245 & 0.2213 & 3.0490 & 3.0052 \\
 & 50   & 0.1229 & 0.2188 & 0.1941 & 0.3872 & 0.1223 & 0.2247 & 3.1224 & 3.0026 \\
 & 100  & 0.1244 & 0.2200 & 0.1853 & 0.3688 & 0.1112 & 0.1976 & 3.0245 & 3.0052 \\
 & FT   & 0.1230 & \textbf{ 0.2085} & 0.1859 & 0.3530 & \textbf{0.1015} & \textbf{0.1939} & 2.9020 & 2.9869 \\
\midrule
\multirow{6}{*}{\begin{tabular}[c]{@{}l@{}}GPT-3.5\\ Turbo-0613-16k\end{tabular}}
 & G    & 0.1540 & \textbf{0.2414} & 0.2688 & 0.5596 & \textbf{0.1378} & \textbf{0.3175} & 3.0408 & 3.0705 \\
 & 10   & 0.1342 & 0.2212 & 0.2462 & \textbf{0.4410} & \textbf{0.1412} & \textbf{0.2531} & 2.9184 & 3.0313 \\
 & 20   & 0.1388 & \textbf{0.2211} & \textbf{0.2465} & \textbf{0.4255} & \textbf{0.1446} & \textbf{0.2394} & 2.9714 & 3.0888 \\
 & 50   & 0.1365 & \textbf{0.2205} & \textbf{0.2437} & 0.4159 & \textbf{0.1356} & \textbf{0.2374} & 2.9918 & 3.0914 \\
 & 100  & 0.1297 & \textbf{0.2207} & \textbf{0.2437} & 0.4092 & \textbf{0.1360} & \textbf{0.2301} & 2.9102 & 3.0470 \\
 & FT   & 0.1162 & 0.1895 & 0.2188 & 0.3341 & 0.0771 & 0.1524 & 2.7143 & 2.9060 \\
\midrule
\multirow{6}{*}{\begin{tabular}[c]{@{}l@{}}GPT-4o\\ 0806-128k\end{tabular}}
 & G    & 0.1992 & 0.2266 & \textbf{0.2739} & \textbf{0.6410} & 0.1224 & 0.2802 & \textbf{3.4612} & 3.4308 \\
 & 10   & 0.1765 & 0.2048 & 0.2455 & 0.4055 & 0.0884 & 0.1963 & \textbf{3.5143} & \textbf{3.5222} \\
 & 20   & 0.1798 & 0.2039 & 0.2453 & 0.4094 & 0.0963 & 0.1830 & \textbf{3.5510} & \textbf{3.5927} \\
 & 50   & 0.1771 & 0.2058 & 0.2433 & \textbf{0.4164} & 0.0971 & 0.1926 & \textbf{3.5592} & \textbf{3.6423} \\
 & 100  & 0.1793 & 0.2036 & 0.2436 & \textbf{0.4120} & 0.0936 & 0.1886 & \textbf{3.5714} & \textbf{3.5614} \\
 & FT   & 0.1821 & 0.1981 & \textbf{0.2372} & \textbf{0.3900} & 0.0713 & 0.1790 & \textbf{3.5673} & \textbf{3.6057} \\
 \bottomrule
\end{tabular}
\captionof{table}{
Evaluation results on the answer generation task of various LLMs, with different context settings (G = Gold Evidence, FT = Full-Text, 10/20/50/100 = Top-k passages) and the metric computed against different ground truths (AE = Answer Evidence Paragraph, FF = Free-Form Answer, GPT-4 FF = GPT-4 rephrased Free-Form Answer). Rouge-L measures lexical overlap; AlignScore measures factual consistency; Prometheus measures answer correctness using an LLM-as-a-judge approach between the generation and the annotated Free-Form Answer or Answer Evidence.}\label{tbl:answer-generation-evaluation}
\end{center}
\end{table}
\clearpage

\section{Answer Generation Error Analysis}\label{sec:appendix-error-analysis}
As outlined in \S\ref{sec:results-answer-generation}, we conducted an error analysis on GPT-3.5's generations. Table~\ref{tbl:appendix-error-class-definition} defines each error class, and Table~\ref{tbl:error-examples} provides an example for each class.
\subsection{Error Classes}
\begin{table}[!htpb]
\small
\begin{tabularx}{\textwidth}{lX}
\toprule
Error Class & Definition \\
\midrule
Evaluation Error & The generated answer is correct; however, at least one of the metrics provides a low score. \\
Partially Correct & The generated answer is correct; however, the free-form answer provides additional details that are not covered by the generation. \\
Reasoning Error & The generation is incorrect. The model fails to arrive at the same conclusion as the free-form answer. \\
Implicit Evidence Only & The generation is incorrect. The evidence only implies the correct answer, making it challenging for the model to infer it. \\
Insufficient Context & The answer is incorrect because further context is required to interpret the evidence correctly (e.g., abbreviations in the context are not resolved, or information established earlier in the paper is missing from the evidence).  \\
Insufficient Free-Form Answer & The author's free-form answer points only to an answer in the paper but does not contain an answer (e.g., "The architecture is depicted in Figures D.3 and D.4").\\
Insufficient Evidence & The highlighted context by the authors does not provide sufficient information to answer the question well. \\
\bottomrule
\end{tabularx}
\caption{Error classes definitions for analyzing the failure modes of the generation models.}
\label{tbl:appendix-error-class-definition}
\end{table}

\clearpage
\subsection{Error Examples}
\begin{table*}[!h]
\begin{adjustbox}{angle=270, width=\textwidth} 
\tiny
\begin{tabularx}{.88\textheight}{@{}p{1.5cm}p{3.5cm}Xp{3.5cm}p{3.5cm}@{}}
\toprule
Error Class & Question & Evidence & Reference Answer & Generated Answer\\ \midrule
Evaluation Error & What does "but train the detector head with $r(\cdot)$ online in a single stage" mean? & To obtain the features for a region b, we apply the VLM pooling layer P(•) on the cropped backbone output features F(I) (see Sec. 3.2 for notations). Because the pooling layer requires fixed-size inputs, e.g. 7x7 for R50 (Radford et al., 2021), we crop and resize the region features with ROI-Align R(•) (He et al., 2017) (see Fig. 2b). Unlike existing works (Gu et al., 2022;Du et al., 2022), we do not crop and resize the RGB image regions and cache their embeddings in a separate offline process, but train the detector head in one stage. This is simpler and more space-efficient. In addition, we do not crop VLM region features with R(•) during training because the backbone features are frozen. & It means that we train the detection coordinates and the class labels for all boxes simlutanously and jointly, similar to Mark-RCNN and related algorithms, as opposed to cropping individual object boxes and feeding their features to a separate model. & "But train the detector head with $r(\cdot)$ online in a single stage" means that the detector head is trained during the same stage as the rest of the training process, rather than in a separate offline process. \\ 
\midrule
Partially Correct & Does the term "aes" refer to the phenomenon that smiling faces are estimated to be older than neutral faces in direct age estimations, or does it also refer to the phenomenon that smiling face group is retrospectively estimated to be younger? & Interestingly, several studies have reported that humans have a counterintuitive bias regarding age. We associate smiling with youth, that is, it is generally believed that when people see a smiling person, they feel that person is younger. Indeed, previous research has provided evidence that individuals with a smile appear younger than those with other facial expressions (Hass et al., 2016;Voelkle et al., 2012). However, contrary to the commonly held association between smiling and youth, Ganel (2015) showed that a smiling face is estimated to be older than a neutral face. This phenomenon, in which smiling faces are evaluated as being older than neutral faces, is called the aging effect of smiling (AES; Ganel \& Goodale, 2021). AES is attributed to wrinkles around the eyes caused by smiling (Ganel, 2015;Ganel \& Goodale, 2021). In contrast, when participants were asked to retrospectively estimate the mean age of several faces (i.e., face group), they estimated that the smiling face group was younger than the neutral face group (Ganel \& Goodale, 2018). These studies indicate that the effect of emotional expressions on age estimation depends on the method of estimation (i.e., directly or retrospectively). & The term AES refers specifically to the phenomenon that a smiling face makes one look older. & The term "aes" refers to both the phenomenon that smiling faces are estimated to be older than neutral faces in direct age estimations and the phenomenon that the smiling face group is retrospectively estimated to be younger. \\ 
\midrule
Reasoning Error & How are the depth and normal maps combined for training the proposal network in phase 1? & There are two possible ways to ensemble geometric cues: (1) Stack the two geometric cues together and train a single object proposal network on these stacked inputs in Phase-I; (2) Train two object proposal networks and extract pseudo boxes separately, then merge them into a single pseudo box pool for Phase-II training. The details of the merging process is described in Appendix A. We conduct ablation studies on these two methods. From Figure 6a, we demonstrate that empirically, ensembling pseudo labels is slightly better than using stacked inputs for Phase-I training. Throughout the paper, we use the pseudo label ensembling for GOOD-Both. & They are combined by merging their pseudo labels. & The depth and normal maps are combined by stacking them together and training a single object proposal network on these stacked inputs in Phase 1. \\
\midrule
Implicit Evidence Only & Does combining RGB, depth, and normals improve the performance of the model for smaller classes? & More comparison of GOOD-Both and GOOD-All.

For GOOD-All, the performance gains in detecting small objects (AR s ) are too small to compensate for the losses in detecting larger objects (AR m and (AR l )), leading to overall inferior performance. & The combination of RGB, depth, and normals does not significantly improve the model's performance for smaller classes. The gains in detecting small objects are too small to offset the losses in detecting larger objects, leading to an overall inferior performance. & The combination of RGB, depth, and normals does not significantly improve the model's performance for smaller classes. The gains in detecting small objects are too small to offset the losses in detecting larger objects, leading to an overall inferior performance.\\ 

\midrule
Insufficient Context & Do the training languages used for the non-baseline models in the cross-lingual experiments have any overlap with German? & In our experiments we make use of the following datasets: The English Nancy Krebs dataset (16h) from the Blizzard challenge 2011 (Wilhelms-Tricarico et al., 2011;King and Karaiskos, 2011); The German dataset of the speaker Karlsson (29h) from the HUI-Audio-Corpus-German (Puchtler et al., 2021); The Greek (4h), Spanish (24h), Finnish (11h), Russian (21h), Hungarian (10h), Dutch (14h) and French (19h) subsets of the CSS10 dataset (Park and Mulc, 2019). In order to investigate the effectiveness of our proposed LAML procedure, we train a Tacotron 2 model and a FastSpeech 2 model on the full Karlsson dataset as a strong baseline. We also train another Tacotron 2 model and another FastSpeech 2 model on speech in 8 languages with one speaker per language (Nancy dataset and CSS10 dataset) and fine-tune those models on a randomly chosen 30 minute subset from the Karlsson dataset. To our surprise, we did not only match, but even outperform the model trained on 29 hours with the model fine-tuned on just 30 minutes in multiple metrics. & The pretraining data does not overlap with German. Only the data used for fine-tuning is in German. & Yes, the training languages used for the non-baseline models in the cross-lingual experiments have overlap with German. The German dataset of the speaker Karlsson is used for training, along with the English Nancy Krebs dataset and subsets of the CSS10 dataset which include Dutch and French. \\
\midrule

Insufficient Evidence & Can the parameters of the BLOSUM matrix be estimated by the model instead of being pre-computed? & In the first ablation study, we investigate the influence of the BLOSUM embeddings by removing them as input to the GRU. Overall, the absence of the BLOSUM embeddings slows down convergence and sometimes make the learning process unstable, but ultimately does not strongly affect accuracy (see Figure 5). 3). The results without the BLOSUM embeddings are shown in pink (MAP) and purple (marginal). & The Blosum matrix is used to serve as a very simple to compute encoding technique to stabilize and accelerate the learning process. With larger datasets it is not necessary to use the Blosum encodings or averaged blosum embedding. & No, the parameters of the BLOSUM matrix cannot be estimated by the model instead of being pre-computed. \\
\midrule

Insufficient Free-Form Answer & What is the architecture of the system being discussed in the paper? & 

\textit{<Omitted for brevity>}
& 
The architecture is depicted in Figures D.3 and D.4 & The architecture is depicted in Figures D.3 and D.4 
\\ 
\bottomrule
\end{tabularx}
\end{adjustbox}
\caption{Examples for each error class.}\label{tbl:error-examples}
\end{table*}

\clearpage
\section{Answer Generation Correlation Analysis}
\subsection{Recall}\label{sec:appendix-correlation-generation-recall}
\begin{figure*}[!b]
    \centering
    \includegraphics[width=0.63\textwidth]{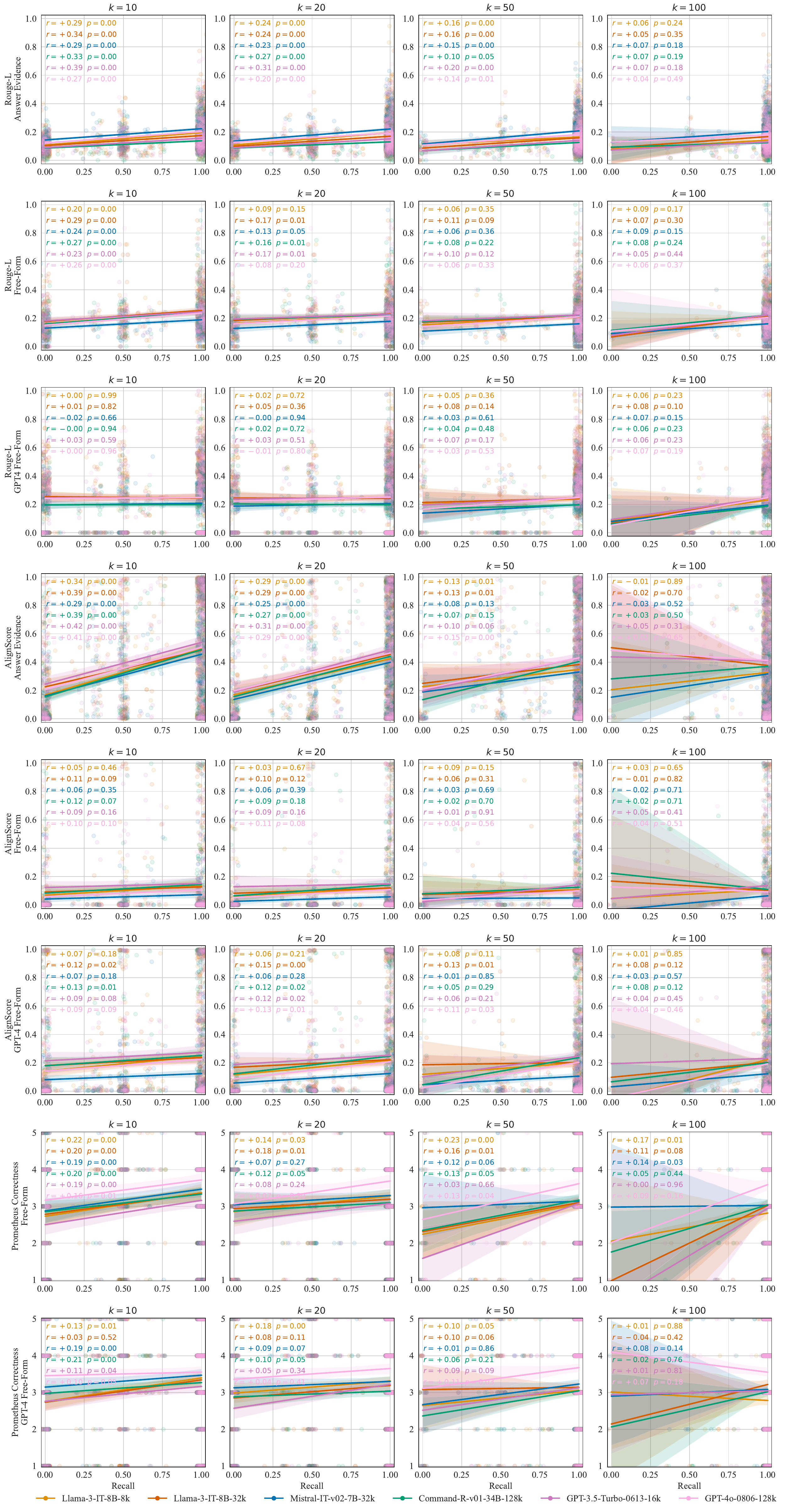}
    \caption{Pearson correlation ($r$) with the corresponding $p$-value between the recall (x-axis) at $k$ (columns) and the answer generation performance (y-axis) according to different metrics (rows). Therefore, each circle represents a single QA pair of a specific model. We added 0.03 x-jitter to the markers to improve visibility.}
    \label{fig:correlation-generation-recall}
\end{figure*}

\begin{multicols*}{2}
Figure~\ref{fig:correlation-generation-recall} visualizes the relationship between the recall of the retrieval model (in this case \texttt{SPLADEv3}) at different cutoffs and the answer generation performance measured by different metrics.
\end{multicols*}

\clearpage
\begin{multicols}{2}
\subsection{Mean Evidence Position}\label{sec:appendix-correlation-evidence-poistion}
\begin{figure*}[!hb]
    \centering
    \includegraphics[width=0.84\textwidth]{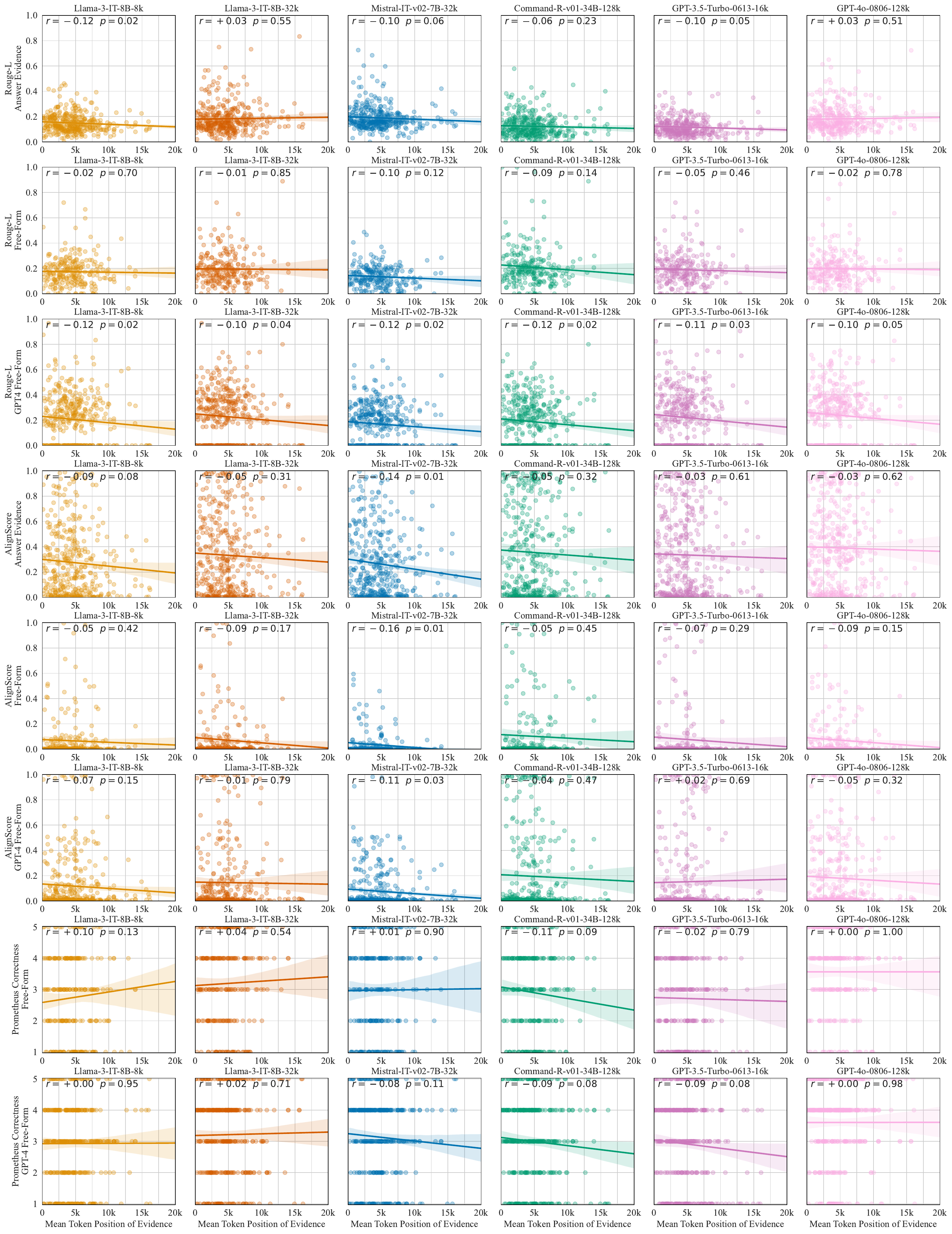}
    \caption{Pearson correlation ($r$) with the corresponding $p$-value between the answer generation evaluation metric (y-axis) and the mean token position of the annotated answer evidence (x-axis).}
    \label{fig:correlation-evidence-poistion}
\end{figure*}

Figure~\ref{fig:correlation-evidence-poistion} visualizes the Pearson correlation between the answer generation metric (Rouge-L, AlignScore, or Prometheus-2 compared to either the answer evidence, the annotated free-form answer or the GPT-4 augmented free-form answer as ground truth) and the mean token position of the answer evidence. All generations are taken from the full-text setting, i.e., where the entire paper text was given as input to the model. To compute the mean token position for each answer evidence, we compute the number of tokens in the paper before the evidence sentence. If a question has multiple answer evidence, we take the average position. We only find a weak relationship that is statistically insignificant in many cases. Nevertheless, some p-values show statistical significance, indicating that for some settings, the generation performance declines when the answer evidence is relatively towards the end of the paper. This finding is also consistent with related work such as \citet{buchmann-etal-2024-attribute}.
\end{multicols}

\clearpage

\begin{multicols}{2}
\section{Answer Generation Similarities}\label{sec:appendix-answer-generation-similarity}
\begin{figure*}[!htbp]
    \centering
    \includegraphics[width=\textwidth]{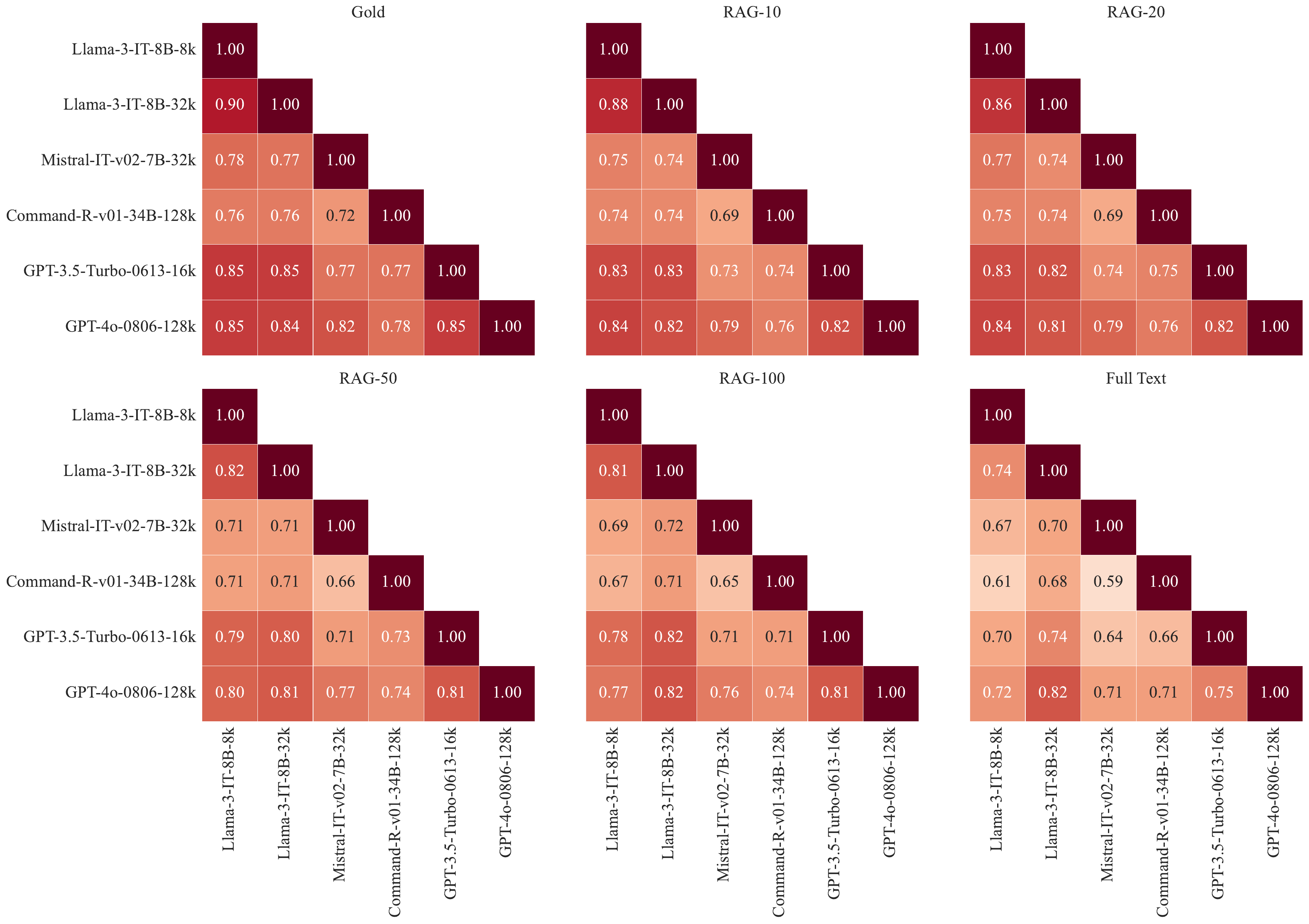}
    \caption{Semantic similarity of the generated answers between models with different context settings.}
    \label{fig:answer-generation-similarity}
\end{figure*}

We compute the average similarity of the generated answers between all models. We embed the generated answers with \texttt{all-MiniLM-L6-v2} and compute the cosine similarity between the generations of the models. Figure~\ref{fig:answer-generation-similarity} visualizes the similarities with the gold and retrieved evidence and full-text settings.
We find that all models produce fairly similar outputs for the gold setting, i.e., where the annotated answer evidence is provided. With increasing retrieved evidence as context (i.e., RAG-10 - RAG-100), the similarity between the model outputs decreases but remains relatively high.
\end{multicols}

\begin{multicols}{2}
\section{Attributable Question Answering}

\begin{Table}
\centering
\begin{tabular}{@{}lcc@{}}
\toprule
 & MRR & Recall@10 \\ \midrule
SPLADEv3 & 0.4536 & 0.6661 \\
GPT-3.5-Turbo-0613-16k & 0.2440 & 0.2762 \\ 
GPT-4o-0806-128k & 0.5429 & 0.5339 \\ 
\bottomrule
\end{tabular}
\captionof{table}{Evidence retrieval scores in the attributable question answering setting.}
\label{tbl:appendix-attributable-qa-evidence-retrieval-results}
\end{Table}

We have considered answer generation based on the top retrieved paragraphs (RAG) or using the full context (\S\ref{sec:experiments-answerability-answer-generation}). In the RAG setup, the answer generation can generally be attributed to the retrieved passages (assuming the model is faithful to the context). However, when using the full text as context, attribution to the passage level is not trivial. Recently, attributable question answering has gained momentum \citep{bohnet-etal-2022-attributed,gao-etal-2023-enabling,malaviya-etal-2024-expertqa}, where in addition to generating an answer, the model is supposed to cite evidence supporting it. Therefore, we also conduct an experiment where the model is conditioned on the full text of the paper and is tasked to "cite" any paragraphs on which the generated answer is based. We prepend an id before each paragraph and include an instruction on how to cite. Specifically, we use the following prompt:

\texttt{
Read the following paper and answer the question. Provide one or several evidence paragraphs that can be used to verify the answer. Give as few paragraphs as possible, but as many that provide evidence to the answer. Your answer must have the following format: "\textless answer\textgreater\ [X] [Y]". In your reply, replace \textless answer\textgreater\ with your answer to the question and add any references in square brackets. Your answer must be followed by the ids of the relevant segments from the document.
Question: \{question\}\\
Paper: \{paper\}\\
Answer:
} 
\\

This setting has the challenge that the model does not provide a ranked list of all paragraphs but an unordered list of what it considers relevant. Therefore, we rank the cited paragraphs in the order in which the LLM generates them.

Table~\ref{tbl:appendix-attributable-qa-evidence-retrieval-results} reports the results of the evidence retrieval with the attributable question answering setup. We find that for \texttt{GPT-3.5}, the scores fall far behind the performance of a dedicated retrieval model (e.g., \texttt{SPLADEv3}). For \texttt{GPT-4o}, the MRR outperforms \texttt{SPLADEv3}, however, the Recall@10 is inferior.

We further investigate the answer generation performance of the attributable QA setup, reporting the results in Table~\ref{tbl:appendix-attributable-qa-answer-generation-results}. Compared with the RAG setting using the top 20 paragraphs retrieved by \texttt{SPLADEv3}, the attributable QA setup performs worse. A RAG setup is also significantly more cost and compute-efficient, particularly considering the long context of papers. Specifically, the average paragraph in PeerQA has 94 tokens, leading to an average of 1880 tokens to encode in the RAG-20 setting. In contrast, on average, a paper has 11723 tokens. Therefore, the full-text setup is 6.24 times more expensive than the RAG-20 setting. 
\end{multicols}

\begin{table*}[h]
\centering
\small
\begin{tabular}{@{}llcccccccc@{}}
\toprule
 &  & \multicolumn{3}{c}{Rouge-L} & \multicolumn{3}{c}{AlignScore} & \multicolumn{2}{c}{Prometheus} \\ 
\cmidrule(lr){3-5}
\cmidrule(lr){6-8}
\cmidrule(lr){9-10}
Model & Ctx. & AE & FF & GPT-4 FF & AE & FF & GPT-4 FF & FF & GPT-4 FF \\
\midrule
\multirow{3}{*}{\begin{tabular}[c]{@{}l@{}}GPT-3.5\\ Turbo-0613-16k\end{tabular}} 
 & 20 & \textbf{0.1388} & \textbf{0.2211} & \textbf{0.2465} & \textbf{0.4255} & \textbf{0.1446} & \textbf{0.2394} & \textbf{2.9714} & \textbf{3.0888} \\
 & FT & 0.1162 & 0.1895 & 0.2188 & 0.3341 & 0.0771 & 0.1524 & 2.7143 & 2.9060 \\
 & FT Cite & 0.1099 & 0.1846 & 0.2057 & 0.2453 & 0.1128 & 0.1564 & 2.4340 & 2.4837 \\
 \midrule
\multirow{3}{*}{\begin{tabular}[c]{@{}l@{}}GPT-4o\\ 0806-128k\end{tabular}}
 & 20 & 0.1798 & \textbf{0.2039} & \textbf{0.2453} & \textbf{0.4094} & 0.0963 & \textbf{0.1830} & 3.5510 & 3.5927 \\
 & FT & \textbf{0.1821} & 0.1981 & 0.2372 & 0.3900 & 0.0713 & 0.1790 & \textbf{3.5673} & \textbf{3.6057} \\
 & FT Cite & 0.1262 & 0.1857 & 0.1602 & 0.2678 & \textbf{0.1177} & 0.1622 & 2.7143 & 2.5614 \\
 \bottomrule
\end{tabular}
\caption{Answer generation scores in the attributable question answering setting ("FT Cite") and two baselines for comparisons. In bold the best performing setup per metric.}
\label{tbl:appendix-attributable-qa-answer-generation-results}
\end{table*}

\end{document}